\documentclass[10pt,twocolumn,letterpaper]{article}

\usepackage[pagenumbers]{cvpr} 

\definecolor{mediumseagreen}{rgb}{0.24, 0.7, 0.44}
\definecolor{malachite}{rgb}{0.04, 0.85, 0.32}
\definecolor{lightgreen}{rgb}{0.56, 0.93, 0.56}
\definecolor{lightercyan}{rgb}{0.92, 1.0, 1.0}
\definecolor{blanchedalmond}{rgb}{1.0, 0.92, 0.8}
\definecolor{mediumpurple}{rgb}{0.58, 0.44, 0.86}
\definecolor{mediumslateblue}{rgb}{0.48, 0.41, 0.93}
\definecolor{brown}{rgb}{0.65, 0.16, 0.16}
\definecolor{burgundy}{rgb}{0.5, 0.0, 0.13}
\definecolor{carnelian}{rgb}{0.7, 0.11, 0.11}
\definecolor{copper}{rgb}{0.72, 0.45, 0.2}
\definecolor{crimson}{rgb}{0.86, 0.08, 0.24}
\definecolor{darkterracotta}{rgb}{0.8, 0.31, 0.36}
\definecolor{dimgray}{rgb}{0.41, 0.41, 0.41}

\newcommand{\gr}[1]{{\color{gray}#1}}
\newcommand{\gn}[1]{{\color{mediumseagreen}#1}}
\newcommand{\pr}[1]{{\color{mediumslateblue}#1}}
\newcommand{\rd}[1]{{\color{darkterracotta}#1}}

\usepackage{amsmath}
\usepackage{graphicx}

\usepackage{float}
\usepackage{booktabs}
\usepackage{multirow}
\usepackage{changepage}
\usepackage{colortbl}
\usepackage{mathabx} 
\usepackage{soul}

\usepackage{amsmath}
\usepackage{algorithm}
\usepackage[noend]{algpseudocode}

\newcommand{\Input}{\State \textbf{Input: }}
\newcommand{\Output}{\State \textbf{Output: }}
\newcommand{\KwDataPrep}{\State \textbf{\textit{Phase 1: Data Preparation}}}
\newcommand{\KwTrain}{\State \textbf{\textit{// Training Loop}}}
\newcommand{\KwVal}{\State \textbf{\textit{// In-loop Validation}}}
\newcommand{\KwValTest}{\State \textbf{\textit{// Full-pipeline Validation}}}
\newcommand{\KwTest}{\State \textbf{\textit{Phase 3: Optimal Experts Combination (after training all stages) and Final Testing}}}
\newcommand{\KwHyperparamSearch}{\State \textbf{\textit{Phase 2: Expert Adapter Search}}}

\definecolor{cvprblue}{rgb}{0.21,0.49,0.74}
\usepackage[pagebackref,breaklinks,colorlinks,allcolors=cvprblue]{hyperref}

\title{MemLoRA: Distilling Expert Adapters for On-Device Memory Systems}

\author{Massimo Bini$^{1,2}$\thanks{Research completed during internship at Samsung R\&D Institute UK.} , Ondrej Bohdal$^{1}$, Umberto Michieli$^{1}$, Zeynep Akata$^{2}$, Mete Ozay$^{1}$, Taha Ceritli$^{1}$ \\ 
{\small$^{1}$Samsung R\&D Institute UK \quad\quad $^{2}$Technical University of Munich,\hspace{0.05cm} Helmholtz Munich,\hspace{0.05cm} MCML}
}

\begin{document}
\maketitle
\begin{abstract}

\noindent
Memory-augmented Large Language Models (LLMs) have demonstrated remarkable consistency during prolonged dialogues by storing relevant memories and incorporating them as context.
Such memory-based personalization is also key in on-device settings that allow users to keep their conversations and data private.
However, memory-augmented systems typically rely on LLMs that are too costly for local on-device deployment. Even though Small Language Models (SLMs) are more suitable for on-device inference than LLMs, they cannot achieve sufficient performance.
Additionally, these LLM-based systems lack native visual capabilities, limiting their applicability in multimodal contexts.
In this paper, we introduce \textit{(i)} \mbox{MemLoRA}, a novel memory system that enables local deployment by equipping SLMs with specialized memory adapters, and \textit{(ii)} its vision extension MemLoRA-V, which integrates small Vision-Language Models (SVLMs) to memory systems, enabling native visual understanding. 
Following knowledge distillation principles, each adapter is trained separately for specific memory operations—knowledge extraction, memory update, and memory-augmented generation. Equipped with memory adapters, small models enable accurate on-device memory operations without cloud dependency.
On text-only operations, MemLoRA outperforms 10× larger baseline models (e.g., Gemma2-27B) and achieves performance comparable to 60× larger models (e.g., GPT-OSS-120B) on the LoCoMo benchmark.
To evaluate visual understanding operations instead, we extend LoCoMo with challenging Visual Question Answering tasks that require direct visual reasoning. On this, our VLM-integrated MemLoRA-V shows massive improvements over caption-based approaches (81.3 vs.~23.7 accuracy) while keeping strong performance in text-based tasks, demonstrating the efficacy of our method in multimodal contexts.

\end{abstract}    
\section{Introduction}
\label{sec:intro}

Recent advancements in Large Language Models (LLMs) and Vision Language Models (VLMs)
have led to their widespread use in conversational Artificial Intelligence (AI) systems, ranging from customer service chatbots to personal assistants and collaborative productivity tools \cite{zhao2023survey,minaee2024large}. VLMs have demonstrated remarkable capabilities in multimodal understanding and generation, making them increasingly integral to human-computer interaction across diverse domains.
However, the effectiveness of VLMs in real-world on-device conversational applications is fundamentally constrained by LLMs' limited context windows \cite{maharana24locomo,chhikara2025mem0buildingproductionreadyai}.
While modern LLMs can process thousands of tokens in a single session, they cannot retain information across multiple conversations or maintain long-term user-specific knowledge. 
This limitation becomes particularly problematic in multi-session scenarios where users expect the system to remember previous interactions, preferences, and contextual details—a critical requirement for delivering truly personalized and coherent conversational experiences.

\begin{figure*}[t]
    \centering \vspace{-0.4cm}
    \includegraphics[width=0.99\textwidth]{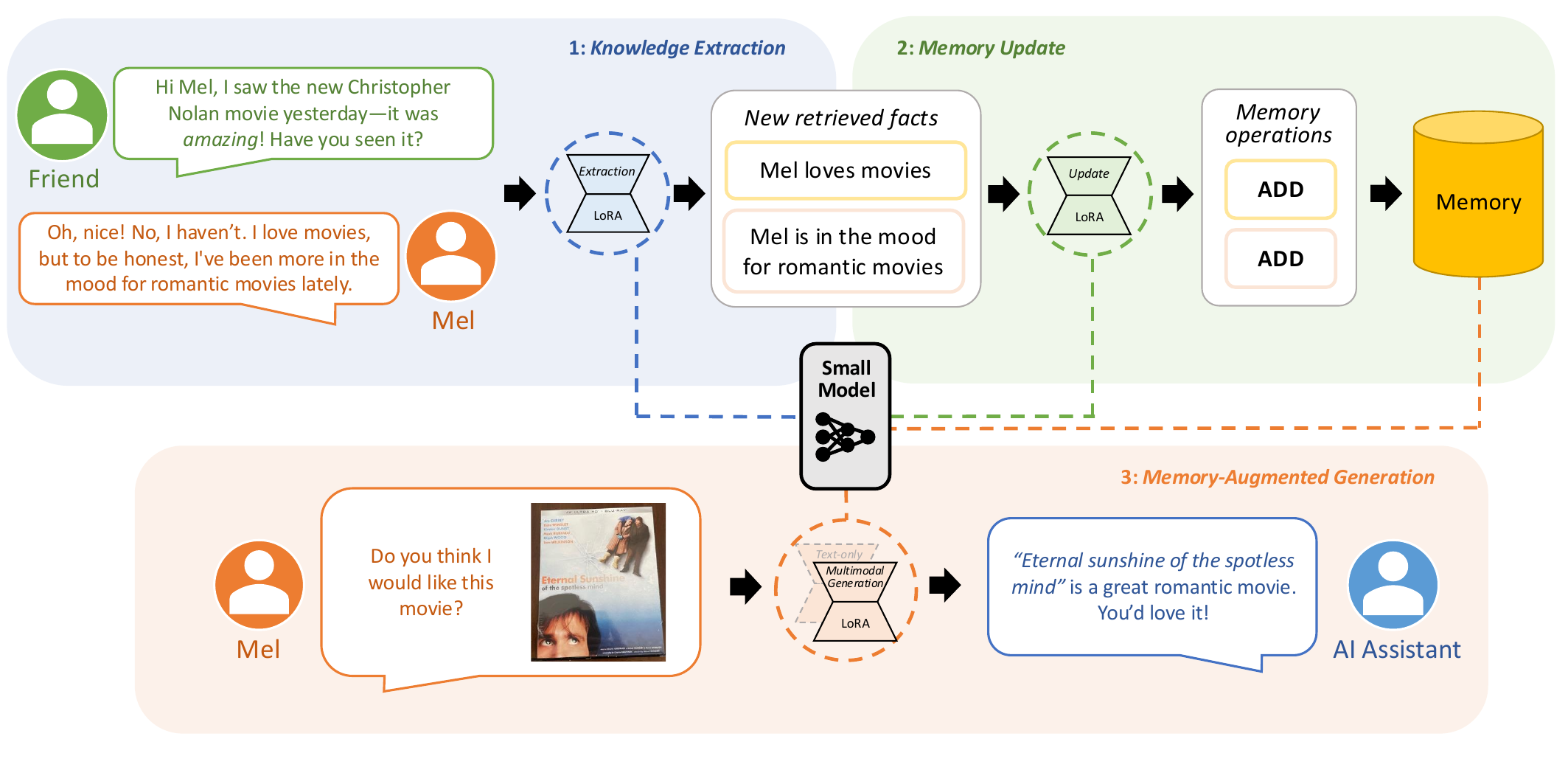}
    \vspace{-0.4cm}
    \caption{\textbf{Overview}.
    We employ specialized LoRA adapters to enable small (vision) language models to perform memory operations for on-device deployment. The base model
    dynamically switches between expert adapters, each trained for a distinct stage: (1) \textit{knowledge extraction}, (2) \textit{memory update}, (3) \textit{memory-augmented generation}. In the last stage, the model can switch between \textit{text-only} and \textit{multimodal} adapter, depending on the input. By specializing each adapter for its specific operation, MemLoRA(-V) 
    achieves performance comparable to models 10-60x larger while enabling efficient local execution without cloud API dependencies.
    }
    \label{fig:overview}
\end{figure*}

To address these challenges, researchers have proposed various memory systems that extend LLMs
with persistent memory capabilities.
Early approaches focused on integrating external memory through differentiable attention mechanisms \cite{Weston2014MemoryN} and retrieval-augmented generation from knowledge bases \cite{NEURIPS2020_6b493230}, establishing foundational paradigms for extending model knowledge beyond immediate context. 
Building on these foundations, recent works have explored sophisticated memory management strategies that mirror human cognitive processes, including temporal decay mechanisms for selective retention \cite{zhong24memorybank}, hierarchical memory systems inspired by the design of operating systems \cite{packer23memgpt}, and knowledge graph representations that track evolving information over time \cite{rasmussen25zep}. 
Contemporary systems have further expanded the role of LLMs
beyond generation, leveraging them as active agents within the memory pipeline itself. Examples include using LLMs 
to automatically extract knowledge and update the memory \cite{chhikara2025mem0buildingproductionreadyai,lee24readagent}, evaluate memory relevance and quality \cite{maharana24locomo}, and dynamically restructure knowledge networks according to emerging patterns \cite{xu2025mem}, thereby transforming memory augmentation from a passive retrieval mechanism into an intelligent, self-improving system.

Despite these advances, current memory systems face significant practical limitations that restrict their deployment and effectiveness. Firstly, these systems fundamentally rely on large, often proprietary, LLMs for core memory operations—including extraction, organization, updating, and retrieval—necessitating continuous API calls to cloud-based services \cite{chhikara2025mem0buildingproductionreadyai,packer23memgpt}. 
This dependency not only introduces latency and cost concerns but also prevents on-device deployment, limiting their applicability in privacy-sensitive contexts, offline scenarios, or resource-constrained environments where cloud connectivity cannot be guaranteed. 
In our work, we tackle this challenge by replacing queries posed to a large-scale model through API, with a small on-device model, equipped with task-specific expert adapters. These adapters are trained via knowledge distillation through teacher answers or ground-truth data.
We provide an overview of the approach and our considered setting in Figure~\ref{fig:overview}.

Secondly, while recent works have begun exploring multimodal capabilities, the handling of visual information remains predominantly text-centric: images are typically converted into textual descriptions through vision-language models before being stored and retrieved \cite{chhikara2025mem0buildingproductionreadyai,lee24readagent}, an approach that inevitably loses fine-grained visual details, spatial relationships, and numerical information embedded in charts or diagrams. This text-first paradigm, though computationally practical, fundamentally constrains the systems' ability to reason directly over visual content, limiting their effectiveness in domains where visual information plays a critical role, such as technical documentation, medical imaging, or design workflows.
Notably, existing benchmarks for evaluating memory systems—such as LoCoMo \cite{maharana24locomo}, which focuses on text-based conversational question answering and event summarization—do not assess multimodal capabilities during inference. Although LoCoMo conversations contain images, the original evaluation relies solely on text-based captions, limiting assessment of native visual understanding. This evaluation gap means that a model's ability to process and reason over visual information directly, rather than through caption intermediaries, remains unmeasured. 

In our work, we address both issues by integrating Vision Language Models (VLMs) in these memory-augmented systems, and by augmenting the LoCoMo benchmark with Visual Questions and Answers (VQA) on the conversational images. 
By doing this, not only we are able to give native visual capabilities to memory-augmented systems, but we are also able to develop our MemLoRA memory system on small VLMs (SVLMs). For these, a novel expert adapter is further introduced to address the VQA task. Such an approach shows how having specialized adapters, one for each operation, can substitute the need for having massive models, and allow for on-device deployment effectively.

We summarize our contributions as follows. 
\textit{(i)} We introduce the challenge of accurate on-device memory systems where small language models are used,  eliminating reliance on cloud-based infrastructure to preserve privacy.
\textit{(ii)} We develop a highly-efficient yet well-performing solution that substantially improves over existing approaches and obtains performance close to that of significantly larger models.
\textit{(iii)} We extend memory systems to incorporate Vision Language Models with native visual capabilities and apply our MemLoRA framework to this multimodal setting through a specialized vision expert adapter.
\textit{(iv)} We augment the LoCoMo benchmark with challenging Visual Question Answering tasks that require direct image access, demonstrating that our approach achieves strong performance with superior efficiency in multimodal contexts.
\section{Related Work}
\label{sec:relatedwork}

\textbf{Memory-Augmented LLMs.}
Memory systems have improved LLMs' capabilities in several applications. Foundational approaches such as Memory Networks \cite{Weston2014MemoryN} and RAG \cite{NEURIPS2020_6b493230} introduced external memory integration and document retrieval.
More sophisticated systems have been inspired by human cognition and operating systems.
Recent innovations like MemoryBank \cite{zhong24memorybank}, MemGPT \cite{packer23memgpt}, Zep \cite{rasmussen25zep}, and Mem0 \cite{chhikara2025mem0buildingproductionreadyai} incorporate hierarchical memory tiers, session management, and self-improving capabilities, while specialized systems like ReadAgent \cite{lee24readagent} and A-Mem \cite{xu2025mem} implement human-inspired organizational principles such as gist memory compression and Zettelkasten-style knowledge networks \cite{zettel1}. 
Many of these approaches rely on agentic frameworks that orchestrate memory operations through iterative LLM queries for tasks such as memory extraction, updating, and retrieval. 
However, such methods require multiple queries to the LLM that are computationally expensive to run and do not prioritize on-device deployment scenarios. In our work, we employ specialized expert adapters on small models to perform memory operations locally, drastically reducing computational demands.

\noindent\\
\textbf{Knowledge Distillation with LLMs.}
Knowledge distillation has evolved into a diverse landscape of techniques aimed at transferring capabilities from powerful teacher models to more efficient student models \cite{xu2024survey,camuffo2025mocha}. 
Generation-aware divergence methods address the limitations of traditional forward Kullback–Leibler Divergence (KLD) by introducing variants such as reverse KLD \cite{gu24minillm} or skew KLD \cite{ko24distillm}, which better handle the challenges of auto-regressive generation while requiring access to the teacher model's internal logits or probability distributions. More recent methods have introduced preference-based frameworks that leverage implicit reward signals \cite{li24dpkd}, advantage functions \cite{gao25adpa}, or pseudo-preference pairs \cite{zhang24plad} to guide the student toward generating outputs that align with the teacher's quality standards \cite{ko25distillm2}. These approaches often necessitate either white-box access to the teacher model's internal states or involve multi-stage optimization procedures. Alternatively, output-only distillation methods that operate solely on generated text sequences represent another direction in the literature \cite{auggpt23, alpaca, zhang24plad, feuer25wildchat}. Such approaches enable distillation from black-box models, proprietary APIs, or any teacher model regardless of architecture, and allow for direct modification of outputs into desired formats or structures before training. In our work, we found that this simpler solution worked well for distilling knowledge in memory systems, and given the practical advantages it offers, we adopted this approach.

\noindent\\
\textbf{Parameter Efficient Finetuning Methods.}
Parameter-efficient finetuning (PEFT) methods have emerged as a powerful approach for adapting large-scale models to specific tasks and domains, while drastically reducing computational requirements compared to full-model finetuning \cite{houlsby2019parameter,lester2021power, lialin23, pmlr-v235-bini24a,ding23}. 
Notable methods in this category are those that inject trainable modules into the model architecture, such as Low-Rank Adaptation (LoRA) \cite{hu2022lora} and its derivations \cite{kopiczko2024vera,liu2024dora,bini2025decouplinganglesstrengthlowrank}. A key advantage of these approaches is that the trained modules can be seamlessly merged and unmerged into the base model weights, eliminating any inference-time latency overhead—a crucial consideration for deployment of large-scale models in production environments—and enabling efficient multi-task setups where a single base model can be dynamically adapted to different tasks or domains simply by swapping the active PEFT module \cite{huang24lorahub}. 
In our work, we leverage these properties
and demonstrate how small (vision) language models with PEFT adapters are able to achieve performance on par with larger counterparts by swapping between expert memory adapters, while significantly increasing efficiency and enabling practical on-device deployment.
\section{Method}
\label{sec:method}
In this section, we present the technical details of MemLoRA, our efficient memory system suitable for on-device deployment. We begin by introducing Mem0 \cite{chhikara2025mem0buildingproductionreadyai}, the memory system we build upon, and describing its core operations (Section~\ref{sec:preliminaries}). We then detail our proposed MemLoRA solution, which replaces the LLM in Mem0 with an SLM and memory adapters through knowledge distillation (Section~\ref{sec:memlora}). Finally, we extend our approach to multimodal settings by incorporating vision understanding capabilities, enabling memory systems to process visual information natively (Section~\ref{sec:visual_memory}).

\subsection{Preliminaries}
\label{sec:preliminaries}
\textbf{Mem0.} 
Mem0 \cite{chhikara2025mem0buildingproductionreadyai} is a memory system enhancing LLM applications with persistent, personalized memory across sessions. Mem0 operates through three main stages:
\\
$\sqbullet$ \textit{Knowledge Extraction.} Given a conversational exchange between a user and an AI assistant, Mem0 uses an extraction prompt to query an LLM $f_{\theta_L}$, parametrized by $\theta_L$.
The extraction prompt guides the LLM to identify relevant knowledge, $\Omega$, consisting of facts, preferences, and contextual information worth storing in memory from the dialogue. 
\\
$\sqbullet$ \textit{Memory Update.} The extracted knowledge $\Omega$ is used to update the memory store $M$. Mem0 queries $f_{\theta_L}$ to determine how new information should be integrated with existing memory $M$—whether to add new entries (\texttt{ADD}), update existing ones (\texttt{UPDATE}), or delete outdated information (\texttt{DELETE}). This ensures the memory remains relevant and consistent over time.
\\
$\sqbullet$ \textit{Memory-Augmented Generation.} During inference, relevant memories $\Omega^\prime$ are retrieved from the memory store $M$ based on semantic similarity to the current query $q$: $\Omega^\prime\leftarrow\text{FindRelatedKnowledge}(M, q)$. 
These memories $\Omega^\prime$ are then provided in the prompt as additional context to $f_{\theta_L}$, enabling it to generate responses that are consistent with past interactions and personalized to the user.

While effective, this approach requires multiple calls to the LLM $f_{\theta_L}$, making it impractical for on-device deployment where computational efficiency, privacy, and offline functionality are critical.

\noindent\\
\textbf{Low-Rank Adaptation (LoRA).}
LoRA \cite{hu2022lora} is a PEFT method to adapt pretrained models by injecting trainable low-rank matrices into specific layers while keeping the original model weights frozen. Given a pretrained weight matrix $W_0 \in \mathbb{R}^{d \times k}$, the LoRA adapter $L$ represents the weight update as the product of two low-rank matrices $(A, B)$:
$$W=W_0+BA,$$
where $B \in \mathbb{R}^{d \times r}$ and $A \in \mathbb{R}^{r \times k}$ with rank $r \ll \min(d, k)$, and $W$ being the updated weights. During training, only the matrices $A$ and $B$ are updated while $W_0$ remains frozen.
LoRA's parameter efficiency and modularity make it ideal for resource-constrained environments. 

Our proposed on-device memory system, MemLoRA, combines Mem0 and LoRA to support multiple task-specific adapters with minimal overhead.

\subsection{Our Method: MemLoRA}
\label{sec:memlora}
MemLoRA addresses the deployment challenges of Mem0 systems by replacing the LLM $f_{\theta_L}$ with a smaller deployable-on-device model $f_{\theta_S}$ parametrized by $\theta_S$ where $|\theta_S|\ll|\theta_L|$ and equipped with multiple specialized memory adapters. Our key insight is that each memory operations—extraction, update, and generation—can be treated as a distinct task amenable to specialized optimization through targeted finetuning.

\noindent\\
\textbf{Memory Adapters.} Given a small language model $f_{\theta_S}$, we employ LoRA to create lightweight expert memory adapters for each memory operation: $L_e$, $L_u$, $L_g$. 
The memory adapters are trained via distilling knowledge from the large model $f_{\theta_L}$.

\noindent\\
\textbf{Knowledge Distillation Signal.} Rather than distilling soft labels or logits from teacher models into memory adapters, we distill from teacher-generated text outputs $y_T\leftarrow f_{\theta_L}(q)$.
We empirically find that training on textual outputs $y_T$ achieves performance close to or exceeding that of teacher models. Such text-based distillation approach offers several practical advantages: \textit{(i)} significant storage reduction compared to saving large logits tensors, \textit{(ii)} flexibility to use student models with different tokenizers than the teacher, and \textit{(iii)} the ability to apply data cleaning and filtering procedures to improve training data quality, and desired outputs, which might differ from base teacher outputs.

\begin{figure*}[t]
    \centering
    \vspace{-0.4cm}
    \includegraphics[width=1.0\textwidth]{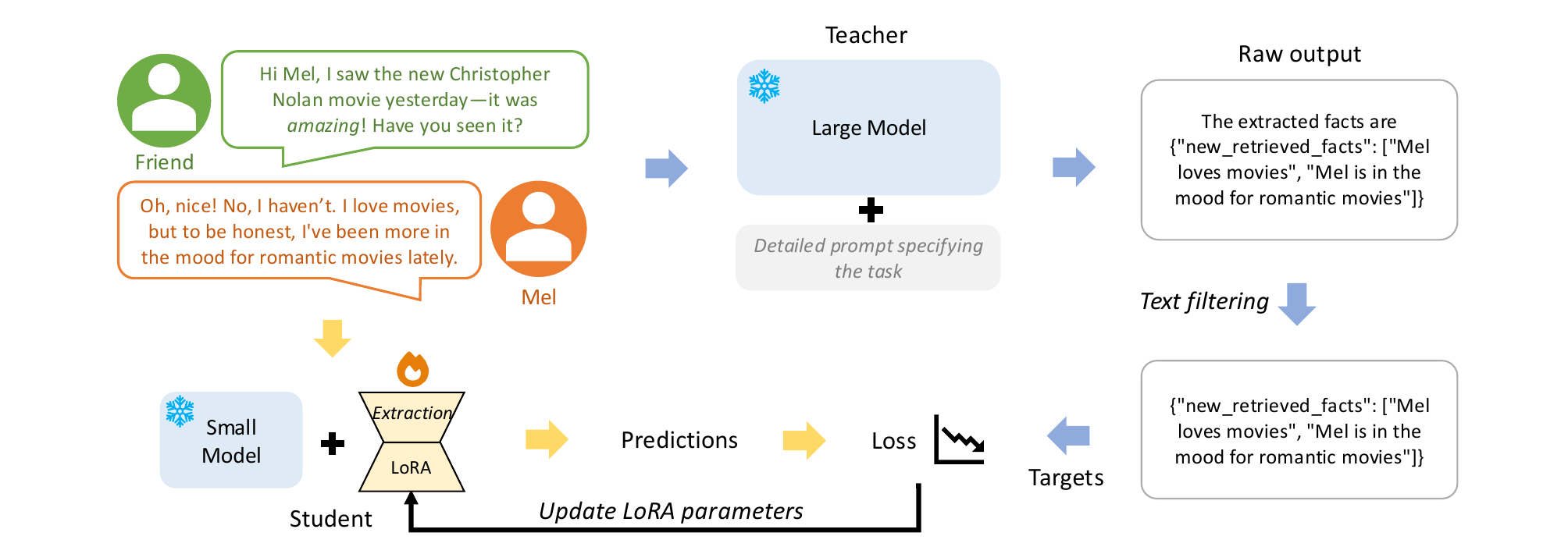}
    \vspace{-0.6cm}
    \caption{\textbf{Training Pipeline (\textit{Extraction} LoRA)}. We first generate outputs for the specific memory-related task via a larger model (teacher). Raw output is further cleaned and used as target for training LoRA parameters of a small model (student).}
    \label{fig:method}
    \vspace{-0.2cm}
\end{figure*}

\noindent\\
\textbf{Data Preparation.} We generate training data by using the teacher model $f_{\theta_{L}}$ on conversational samples from the LoCoMo dataset, then applying operation-specific processing (detailed examples are provided in Section~\ref{sec:data_prep}):
\noindent\\
$\sqbullet$ \textit{Extraction Adapter.} We train on teacher-generated extractions. We do simple cleaning by removing the ``thinking process" of the model, and keeping the minimal \texttt{json} form output.
\noindent\\
$\sqbullet$ \textit{Update Adapter.} We observe the teacher model predicts unnecessary \texttt{NONE} (i.e. no action) operations for previously retrieved memories rather than focusing solely on newly extracted knowledge. 
In addition to standard cleaning as before, we filter the training data to process only updates related to new extractions, improving efficiency and focus.
\noindent\\
$\sqbullet$ \textit{Generation Adapter.} We leverage teacher-generated memory banks for contextual input to the student model, however we train directly on ground-truth responses from the LoCoMo benchmark rather than teacher-generated outputs. This ensures that the generation expert learns from optimal rather than suboptimal examples, being teacher model accuracies around 40-50\%.

\noindent\\
\textbf{Training Pipeline.} The detailed pipeline is provided in Section~\ref{sec:training_pipeline}, Algorithm \ref{alg:training_pipeline}.
For each expert adapter $L_e, L_u, L_g$, we: \textit{(i)} generate or prepare training data using the appropriate source (teacher outputs or ground truth), and apply operation-specific cleaning and filtering procedures, \textit{(ii)} independently train each expert adapter using standard next-token prediction with cross-entropy loss, enabling specialization without interference, and \textit{(iii)} integrate and test the full pipeline with trained adapters.

This process yields three expert adapters: an \textit{extraction expert} $L_e$ for identifying relevant information from conversations, an \textit{update expert} $L_u$ for memory management decisions, and a \textit{generation expert} $L_g$ for producing memory-augmented responses.
An illustration of the training pipeline for one adapter is provided in Figure~\ref{fig:method}.

\noindent\\
\textbf{Inference Pipeline.} During deployment, MemLoRA operates identically to Mem0 but dynamically loads the appropriate expert adapter at each stage. The base SLM, $f_{\theta_S}$, switches between memory adapters as needed—\textit{extraction expert} $L_e$ for knowledge identification, \textit{update expert} $L_u$ for memory modifications, and \textit{generation expert} $L_g$ for response creation—maintaining the same three-stage pipeline while drastically reducing computational requirements and enabling fully-local execution.

\subsection{Native Visual Understanding Capabilities}
\label{sec:visual_memory}

\begin{figure}[t]
    \vspace{-0.15cm}
    \centering
    \includegraphics[width=\columnwidth]{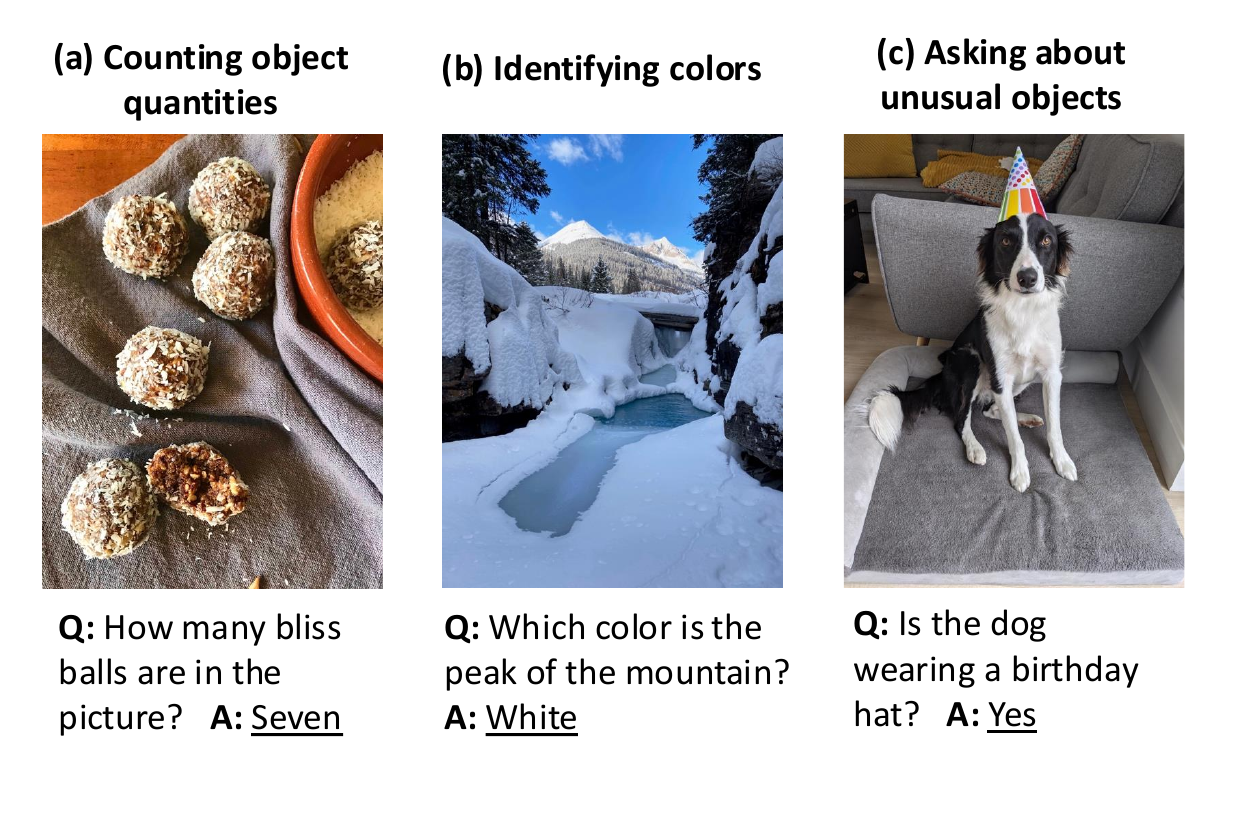}
    \vspace{-1.1cm}
    \caption{\textbf{Our augmentation of LoCoMo} includes challenging VQA tasks about (a) counting object quantities, (b) identifying colors, and (c) asking about unusual objects.}
    \label{fig:benchmark}
    \vspace{-0.3cm}
\end{figure}

While language-based memory systems have proven effective for dialogue, real-world conversations often involve visual elements—shared images, screenshots, or visual references. Previous memory systems, including the original Mem0, processed images during the knowledge extraction phase, by using a BLIP captioning model \cite{li2022blip} to extract general information about images in the conversation.
However, this caption-based approach introduces two critical limitations: \textit{(i)} once images are captioned during extraction, any information not captured in the caption is permanently lost, preventing later queries about visual details, and \textit{(ii)} querying images on-the-fly is not natively supported, requiring a separate model to extract information from them.

\noindent\\
\textbf{Mem0-V.}  To address these limitations, we extend Mem0 to use Vision Language Models (VLMs). By replacing the earlier foundation model with a VLM, Mem0-V enables \textit{(i)} native knowledge extraction without requiring a separate image processor, and \textit{(ii)} direct image processing in queries posed to the system, while keeping the remaining pipeline the same. This allows the system to access rich visual information throughout all memory operations rather than relying solely on pre-generated captions.

\noindent\\
\textbf{MemLoRA-V.} We extend our efficient solution analogously by replacing the base SLM with a Small Vision Language Model (SVLM), yielding MemLoRA-V with native visual capabilities for on-device deployment. To support visual understanding, we introduce a fourth expert adapter specifically trained on Visual Question Answering (VQA) tasks using images from the LoCoMo dataset \cite{maharana24locomo}. 
Following our distillation approach for language experts, we train this vision expert $L_g^V$ on output data generated by a larger vision-language teacher model. When \mbox{MemLoRA-V} receives a query about an image, it activates the vision expert adapter $L_g^V$ rather than the language-based one $L_g$, leveraging specialized visual reasoning capabilities to process the image effectively.

\noindent\\
\textbf{LoCoMo VQA Augmentation.} To evaluate these native image understanding capabilities, we recognize that the original LoCoMo questions are insufficient—they can often be answered using captions alone or do not require visual reasoning at all. Therefore, we create a novel VQA benchmark that augments LoCoMo with challenging visual questions about images already present in the dataset.
To automate the creation of these challenging questions and ground-truth answers, we employ InternVL3-78B \cite{zhu2025internvl3}, one of the strongest open-source VLMs available at the time of development. We design questions to be ``challenging" by instructing the model to generate queries following three types: (a) counting object quantities, (b) identifying colors of specific image regions, and (c) asking about unusual objects in the scene, as illustrated in Figure~\ref{fig:benchmark}. These question types were selected after evaluating eight alternatives, where a validator model (InternVL3-2B) attempted to answer each type. The three types that resulted in the highest error rates were chosen to construct our benchmark, ensuring the task requires genuine visual reasoning. Further details and examples are provided in Section~\ref{sec:supp_vqa}.

While most VQA benchmarks use open-ended questions \cite{duan24vlmbench,lee24vlmbench}, this format requires resource-intensive LLM-as-a-judge approaches to reliably assess answer correctness \cite{zhang25autovqa}. 
To enable efficient evaluation, we design our questions to be easily assessable, by basing the evaluation on single-word answers. Specifically, we instruct InternVL3-78B to generate questions answerable with a single word, and structure responses accordingly as: \texttt{"answer": "<one-word-answer>"} and \texttt{"reason": "<explanation>"}.
The single-word answer enables evaluation using word similarity metrics, eliminating the ambiguity inherent in free-form responses. The reason field accommodates VLMs' natural tendency to explain their reasoning, making the format more aligned with how these models generate outputs. During supervised finetuning of the SVLM vision expert adapter, we leverage both fields to provide a richer training signal, which improves the expert's visual reasoning capabilities beyond what the answer alone would provide.

In summary, we introduce three key contributions for multimodal memory systems: Mem0-V, which extends the original Mem0 memory system with native VLM capabilities; MemLoRA-V, our efficient on-device variant with a specialized vision expert adapter; and a novel VQA benchmark augmentation for LoCoMo that enables efficient evaluation of visual reasoning in memory-augmented systems.
\section{Experiments}
\label{sec:experiments}

In this section, we evaluate our proposed method MemLoRA on memory-augmented dialogue and multimodal conversation understanding tasks, comparing its performance against Mem0 baselines utilizing models of varying sizes. We demonstrate that MemLoRA achieves competitive performance with significantly larger models while providing massive improvements in computational efficiency for on-device deployment.

\subsection{Experimental Setup}

To evaluate MemLoRA's performance, we integrate it within the Mem0 memory system \cite{chhikara2025mem0buildingproductionreadyai} and follow the same evaluation setup. Specifically, we utilize the Question Answering (QA) task of the LoCoMo benchmark \cite{maharana24locomo} to assess long-term conversational memory in AI agents. This benchmark features 10 extended, multi-session dialogues, each with hundreds of turns, and includes questions categorized as single-hop, multi-hop, temporal, and open-domain.
In the context of Mem0 and our method, the evaluation measures the ability of different LLMs to \textit{(i)} extract useful knowledge from conversational data, \textit{(ii)} update memory storage with necessary information, and \textit{(iii)} correctly utilize the retrieved memory context.
In our VLM-integrated benchmark, we further introduce a new VQA task, where the model is asked three challenging types of questions on each image present in the conversation, evaluating the model's performance in assisting with visual data.

\noindent\\
\textbf{Data Split.} Given the necessity of training our expert adapters to perform the different memory operations, we split the LoCoMo dataset into training, validation, and test sets, following an approximate 70-10-20\% split respectively. To prevent data leakage and ensure valid evaluation, we keep entire conversations together within each split. All results reported in our benchmark tables are computed on the held-out 20\% test split, on which no hyperparameter tuning was performed.

\noindent\\
\textbf{Metrics.} 
The experimental setup measures performance using two metrics: \textit{(i)} a composite score \textit{L} that aggregates surface text-matching metrics (ROUGE-1 \cite{lin04rouge} and METEOR \cite{BanerjeeL05meteor}) with semantic metrics (BERTScore-F1 \cite{Zhang20bertscore} and SentenceBERT \cite{reimers2019sentencebert}); and \textit{(ii)} an LLM-as-a-Judge score \textit{J} \cite{gu2024surveyllmasajudge} for in-depth reasoning evaluation.
\textit{L} evaluates similarity to ground-truth answers and is computationally efficient. In contrast, \textit{J} serves as our primary factual accuracy metric, as LLM-based evaluation has proven more effective for this purpose \cite{janiak25illusion, chhikara2025mem0buildingproductionreadyai}, though it requires significantly greater computational resources.
To allow for reproducibility over time, we do not use API-based models as the evaluator model (Judge), but rather GPT-OSS-120B \cite{gpt120b}, being one of the most capable open-source models that can fit on a single A100-80GB-GPU.
The metric for the VQA task, denoted as \textit{V}, is the average matching between the predicted one-word answers from the tested model against the ones generated by InternVL3-78B \cite{zhu2025internvl3} when creating the dataset.

\subsection{Benchmark Results}

We compare our MemLoRA approach with Mem0, which are both powered by open-source locally-downloaded models for fair comparison and reproducibility.

\noindent\\
\textbf{Language-only Memory Systems.} In the setup with language models utilization, we test Mem0 with different baseline models: two large language models (LLMs), namely Gemma2-27B \cite{gemmateam2024gemma2improvingopen} and GPT-OSS-120B \cite{gpt120b}, and two small language models (SLMs), namely Qwen2.5-1.5B \cite{qwen2025qwen25technicalreport} and Gemma2-2B \cite{gemmateam2024gemma2improvingopen}.
We test our MemLoRA by equipping memory adapters
to the two SLMs, powered via knowledge distillation from teachers' data.
Table \ref{tab:llmbench} presents these results, showing MemLoRA surpasses the Gemma2-27B baseline by a significant margin on three student-teacher combinations out of four. Here, the leading MemLoRA variant, with Gemma2-2B finetuned using Gemma2-27B generated data, achieves a \textit{J} score of 47.2, much larger than 39.1 of Gemma2-27B, and comparable to 48.9 of GPT-OSS-120B. 

\begin{table}[t]
\centering
\vspace{-0.1cm}
\caption{\textbf{Comparison of \pr{MemLoRA} against Mem0 on LoCoMo.} Evaluation done in terms of composite score \textit{L}, and LLM-as-a-judge score \textit{J}. $\Delta$\textit{J}$^{\text{\textit{base}}}$ measures the relative improvement with respect to the base SLM. By equipping 1.5B/2B SLMs with memory adapters, MemLoRA surpasses 27B models, reaching comparable results to 120B ones.}
\vspace{-0.25cm}
\label{tab:llmbench}
\begin{tabular}{ccccc}
\toprule
\textit{LLM}& \hspace{-0.3cm}\textit{KD teacher} & \textit{L} & \textit{J} \hspace{-0.1cm}&\hspace{-0.4cm} \textit{$\Delta$\textit{J}$^{\text{\textit{base}}}$}\hspace{-0.4cm} \\
\midrule
\hspace{-0.2cm}Gemma2-27B \hspace{-0.3cm}& \hspace{-0.3cm}-&\hspace{-0.3cm} 38.6 &\hspace{-0.2cm} 39.1 &\hspace{-0.3cm}-\\
\hspace{-0.2cm}GPT-OSS-120B \hspace{-0.3cm}&\hspace{-0.3cm}- & \hspace{-0.2cm}38.9 & \hspace{-0.1cm}48.9&\hspace{-0.3cm}- \\
\hline
\hspace{-0.2cm}Qwen2.5-1.5B &\hspace{-0.3cm}-& \hspace{-0.2cm}30.5 & \hspace{-0.1cm}29.6 &\hspace{-0.3cm}-\\
\pr{+Exp (ours)} \hspace{-0.3cm}& \hspace{-0.3cm} \pr{\scalebox{0.99}{Gemma2-27B}} \hspace{-0.3cm}& \hspace{-0.3cm} 37.3 & 36.9 \hspace{-0.1cm}&\hspace{-0.3cm} +25\% \hspace{-0.4cm}\\
\pr{+Exp (ours)} \hspace{-0.3cm}& \hspace{-0.3cm} \pr{\scalebox{0.99}{GPT-OSS-120B}} \hspace{-0.3cm}& \hspace{-0.3cm} 38.4 & 42.1 \hspace{-0.1cm}&\hspace{-0.3cm} +42\% \hspace{-0.4cm}\\
\hline
\hspace{-0.2cm}Gemma2-2B  & \hspace{-0.3cm}-& \hspace{-0.2cm}29.1 & \hspace{-0.1cm}24.9 &\hspace{-0.3cm}- \\
\pr{+Exp (ours)}  \hspace{-0.3cm}& \hspace{-0.3cm}\pr{\scalebox{0.99}{Gemma2-27B}} \hspace{-0.3cm}& \hspace{-0.3cm} \textbf{44.5} & \textbf{47.2} \hspace{-0.1cm}&\hspace{-0.3cm} +90\% \hspace{-0.4cm}\\
\pr{+Exp (ours)}  \hspace{-0.3cm}& \hspace{-0.3cm}\pr{\scalebox{0.99}{GPT-OSS-120B}} \hspace{-0.3cm}& \hspace{-0.3cm} \underline{42.7} & \underline{44.6} \hspace{-0.1cm}&\hspace{-0.3cm} +79\% \hspace{-0.4cm}\\
\bottomrule
\vspace{-0.7cm}
\end{tabular}
\end{table}

\noindent\\
\textbf{Vision-Language-integrated Memory Systems.} In our novel vision-language integration within the memory system, we compare our VLM-integrated Mem0-V with our VLM-integrated MemLoRA-V. We evaluate these models in both the standard QA task from LoCoMo, and on our newly introduced VQA task. As small VLMs, we use InternVL3-1B and InternVL3-2B \cite{zhu2025internvl3} equipped with one adapter trained on text-only QA as before, and a new adapter trained on VQAs with images from the training set. To highlight the abilities of VLM-integrated systems, we also compare these methods with text-only Mem0 vision baselines. For this case, we adapt the VQA tasks to use text coming from BLIP \cite{li2022blip} captions, as utilized by Mem0 in the extraction stage. 

Table \ref{tab:vlmbench} presents these results. Interestingly, in the text-only QA task, our MemLoRA-V  applied on InternVL3-2B and InternVL3-1B, surpasses larger text-only models such as Gemma2-27B.
At the same time, in the VQA task, we observe significant improvements for these VLMs with dedicated adapters, increasing \textit{V} score from 50.0 to 69.4 for InternVL3-1B, and from 70.8 to 81.3 for InternVL3-2B. In contrast, Mem0 that only uses text-based BLIP captions performs significantly worse than these VLM-integrated variants, reaching a highest value of 23.7, showing one limitation of language-only systems.

\begin{table}[t]
\centering
\vspace{-0.1cm}
\caption{\textbf{Comparison of \pr{MemLoRA-V} and Mem0-V, as well as the original Mem0, on LoCoMo benchmark and newly introduced VQA task.} Evaluation done in terms of composite score \textit{L}, LLM-as-a-judge score \textit{J}, and accuracy in our VQA task (\textit{V}). G-27 stands for Gemma2-27B, IVL3-78B stands for InternVL3-78B. Notice how by training specialized adapters on both tasks, Mem0-V is able to achieve strong performance in both, while keeping resource utilization low. *LLM-based Mem0 baselines, utilize BLIP extracted captions as contextual information on the images.}
\vspace{-0.25cm}
\label{tab:vlmbench}
\begin{tabular}{ccccc}
\toprule
\hspace{-0.2cm}\textit{LLM/VLM}& \hspace{-0.2cm}\textit{KD teacher} & \textit{L} & \textit{J} & \hspace{-0.1cm}\textit{V}\hspace{-0.1cm} \\
\midrule
\hspace{-0.2cm}Gemma2-27B \hspace{-0.2cm}&\hspace{-0.25cm}-& 38.6 & 39.1 & \hspace{-0.1cm}23.7*\hspace{-0.1cm} \\
\hspace{-0.2cm}GPT-OSS-120B \hspace{-0.2cm}&\hspace{-0.25cm}-& 38.9 & 48.9 & \hspace{-0.1cm}22.0*\hspace{-0.1cm} \\
\hline
\hspace{-0.2cm}InternVL3-1B\hspace{-0.2cm} &\hspace{-0.25cm}- & 13.7 & 9.0 & \hspace{-0.1cm}50.0\hspace{-0.1cm} \\
\hspace{-0.2cm}\pr{+Exp (ours)} & \hspace{-0.25cm}\pr{\scalebox{0.98}{G-27B\textbackslash IVL3-78B}}\hspace{-0.15cm} &29.1 & 20.2 & \hspace{-0.1cm}\underline{69.4}\hspace{-0.1cm} \\
\hline
\hspace{-0.2cm}InternVL3-2B\hspace{-0.25cm}&\hspace{-0.25cm}-& \underline{32.2} & \underline{27.0} & \hspace{-0.1cm}70.8\hspace{-0.1cm} \\
\hspace{-0.2cm}\pr{+Exp (ours)} & \hspace{-0.25cm}\pr{\scalebox{0.98}{G-27B\textbackslash IVL3-78B}}\hspace{-0.15cm} & \textbf{44.6} & \textbf{40.3}& \hspace{-0.1cm}\textbf{81.3}\hspace{-0.1cm} \\
\bottomrule
\vspace{-0.5cm}
\end{tabular}
\end{table}

\subsection{Efficiency Measures}
One main advantage of MemLoRA is its efficient deployment capability. Specifically, compared to API-based memory systems that rely on cloud-hosted large language models, MemLoRA enables fully local execution with significantly reduced computational requirements, lower latency, and no dependency on network connectivity. By replacing a single large LLM with specialized lightweight adapters on small language models, our solution drastically reduces memory footprint, and inference time—critical factors for on-device deployment scenarios such as mobile applications, edge devices, and privacy-sensitive environments. 

\begin{table}[t]
\centering
\vspace{0.1cm}
\caption{\textbf{Comparison of \pr{MemLoRA} and Mem0 in terms of efficiency.} Under the same computational resources, MemLoRA requires 10-20× smaller memory and delivers 10-20× faster responses with respect to LLM-powered Mem0, while achieving comparable performance}
\vspace{-0.25cm}
\label{tab:efficiency}
\begin{tabular}{ccccc}
\toprule
\textit{LLM} & \hspace{-0.2cm}\textit{size(GB)}\hspace{-0.2cm} & \textit{tok/s}$\uparrow$ & \textit{tok/ans}$\downarrow$ & \textit{s/ans}$\downarrow$ \\ 
\midrule
\hspace{-0.cm}Gemma2-27B & 50.71 &9.2&97.63&10.66 \\
\hspace{-0.cm}GPT-OSS-120B & 60.77 & 11.4&209.91 &22.82 \\
\hline
\hspace{-0.cm}Qwen2.5-1.5B & 2.88 & 71.0&54.74&0.77 \\
\hspace{-0.cm}\pr{+Exp (ours)} & 2.92 &71.0 & \textbf{45.26}&\textbf{0.64} \\ 
\hline
\hspace{-0.cm}Gemma2-2B & 4.87 & 47.4&33.13&0.70 \\
\hspace{-0.cm}\pr{+Exp (ours)} & 4.92 & 47.4&\textbf{32.73}&\textbf{0.69} \\
\bottomrule
\vspace{-0.6cm}
\end{tabular}
\end{table}

In Table~\ref{tab:efficiency} we report efficiency measures of MemLoRA compared with Mem0 baselines utilizing LLMs of different sizes. Specifically, we report model sizes and operational measures such as tokens per second (\textit{tok/s}), tokens per LLM answer (\textit{tok/ans}), and seconds per LLM answer (\textit{s/ans}). These latter measures are obtained by averaging over all three memory stages of \textit{knowledge extraction}, \textit{memory update}, and \textit{memory-augmented generation}, while operating to a portion of the LoCoMo benchmark. We calculate these metrics by averaging over multiple runs, maintaining the setup unaltered.
In standard Mem0, deploying larger models on-device yields strong performance but results in 10-30x slower inference, whereas using smaller models improves efficiency but compromises accuracy, highlighting a fundamental performance-efficiency trade-off. \mbox{MemLoRA} bridges this gap, matching the performance of significantly larger models while retaining the efficiency of small models through task-specialized expert adapters.
Furthermore, compared to base SLMs, by formatting their output to match the memory system usage, we are able to reduce the number of tokens per answer (quantitative comparisons are reported in Section~\ref{sec:supp_outtokens}, Table~\ref{tab:efficiency4}), further reducing the operational time of the memory system.

\subsection{Ablations}

We validate our design choices via two comprehensive ablations: \textit{(i)} we study the contribution of each memory adapter at different stages of the memory pipeline; and \textit{(ii)} we study the impact of student model size on overall performance.

\begin{table}[b]
\centering
\caption{\textbf{Ablation of MemLoRA adapters (\pr{+Exp}) for each operation}, comparing Gemma2-2B (G-2B) equipped with experts against its teacher Gemma2-27B (\rd{G-27B}). In \textit{extraction} and \textit{update} stages, MemLoRA shows stronger performance than the teacher, 
being trained on filtered teacher-generated data. 
In \textit{generation}, specialization on the QA task yields the largest gain, with the expert largely surpassing the teacher model (47.2 vs.\ 39.1).}
\label{tab:singled}
\vspace{-0.2cm}
\setlength{\tabcolsep}{5pt}
\begin{tabular}{cccccc}
\toprule
\textit{extraction} & \textit{update} & \hspace{-0.1cm}\textit{generation}\hspace{-0.1cm} &\hspace{-0.1cm} \textit{L} & \textit{J} & \hspace{-0.2cm}$\Delta$\textit{J}$^{\text{\textit{prev}}}$\hspace{-1cm}\\
\midrule
G-2B&G-2B&G-2B & 29.1 & 24.9 &\hspace{-0.8cm}-\hspace{-0.6cm} \\
\hline
\rd{G-27B}&G-2B&G-2B & 32.7 & 30.9 & \hspace{-0.2cm}+24\%\hspace{-0.6cm}\\
\rd{G-27B}&\rd{G-27B}&G-2B & 34.7 & 34.8 & \hspace{-0.1cm}\textbf{+13\%}\hspace{-0.6cm}\\
\rd{G-27B}&\rd{G-27B}&\rd{G-27B} & 38.6 & 39.1 & \hspace{-0.1cm}+12\% \hspace{-0.6cm}\\
\hline
\hspace{-0.1cm}G-2B\pr{\scalebox{0.99}{+Exp}}&G-2B&G-2B& 32.9 & 32.2 & \hspace{-0.1cm}\textbf{+29\%}\hspace{-0.6cm}\\
\hspace{-0.1cm}G-2B\pr{\scalebox{0.99}{+Exp}}\hspace{-0.3cm}&\hspace{-0.1cm}G-2B\pr{\scalebox{0.99}{+Exp}}&G-2B& 35.1 & 35.6 & \hspace{-0.1cm}+11\%\hspace{-0.6cm}\\
\hspace{-0.1cm}G-2B\pr{\scalebox{0.99}{+Exp}}\hspace{-0.3cm}&\hspace{-0.1cm}G-2B\pr{\scalebox{0.99}{+Exp}}\hspace{-0.4cm}&\hspace{-0.4cm}G-2B\pr{\scalebox{0.99}{+Exp}}\hspace{-0.4cm} & 44.5 & 47.2 & \hspace{-0.1cm}\textbf{+33\%}\hspace{-0.6cm}\\
\bottomrule
\end{tabular}
\end{table}

\noindent\\
\textbf{Per-stage Incremental Performance.} To isolate the contribution of each expert adapter, in Table \ref{tab:singled} we report a stage-wise ablation study evaluating performance improvements at each memory operation. We measure the impact of our specialized adapters for \textit{knowledge extraction}, \textit{memory update}, and \textit{memory-augmented generation} independently. In the extraction and update stages, MemLoRA demonstrates strong performance even when trained on data generated by Gemma2-27B, with the trained experts showing notable robustness across different conversational contexts. Most significantly, in the generation stage, specializing the adapter directly on the QA task yields the largest performance gain, with our generation expert achieving a \textit{J} score of 47.2 compared to the teacher model's 39.1. This substantial improvement—surpassing the teacher by 8.1 points—demonstrates that task-specific specialization through dedicated memory adapters can not only match but exceed the capabilities of general-purpose larger models, particularly when trained on high-quality ground-truth data.

\noindent\\
\textbf{Student's Performance at Different Scales.} To understand how student model capacity affects our approach, we evaluate MemLoRA across multiple model sizes in Table~\ref{tab:student}, ranging from compact models for resource-constrained devices to moderately-sized alternatives. 
Our results reveal that increasing the student model size initially yields substantial performance improvements, with gains progressively decreasing as student models grow larger.

\begin{table}[t]
\centering
\caption{\textbf{Ablation evaluating the effect of MemLoRA at different students' scales.} As expected, we find that the smallest models lead to the largest improvements, while we see diminishing improvements as the students' size increases.}
\label{tab:student}
\vspace{-0.3cm}
\begin{tabular}{ccccc}
\toprule
\textit{LLM}& \textit{KD teacher} & \textit{L} & \textit{J} & \textit{$\Delta$\textit{J}$^{\text{\textit{base}}}$} \\
\midrule
Qwen2.5-0.5B &-& 19.5 & 11.2 & - \\
\pr{+Exp (ours)} & \pr{\scalebox{0.99}{Gemma2-27B}} & 28.1 & 26.6 & \hspace{-0.1cm}+138\% \\
\hline
Qwen2.5-1.5B &-& 30.5 & 29.6 & -\\
\pr{+Exp (ours)} & \pr{\scalebox{0.99}{Gemma2-27B}} & 37.3 &  36.9 & \hspace{-0.1cm}+25\% \\
\hline
Qwen2.5-3B &-& 39.9 & 35.6 &- \\
\pr{+Exp (ours)}& \pr{\scalebox{0.99}{Gemma2-27B}} & 42.3 & 42.1 & \hspace{-0.1cm}+18\% \\
\bottomrule
\vspace{-0.7cm}
\end{tabular}
\end{table}

\section{Conclusions}
\label{sec:conclustions}

In this work, we introduced MemLoRA, a novel memory system enabling efficient on-device deployment of memory-augmented systems through specialized memory adapters on small models. By treating each memory operation as a distinct task, we demonstrate that lightweight adapters achieve performance comparable to models 10-60× larger while drastically reducing computational requirements and enabling local execution.
Our evaluation on the LoCoMo benchmark validates this approach. 
Our ablation studies reveal that expert adapters consistently surpass teacher models, and that performance exhibits diminishing returns with increasing student model size. 
We extend our approach to multimodal settings with \mbox{MemLoRA-V}, featuring native visual understanding via a specialized vision expert adapter. To assess this, we enhanced LoCoMo with challenging VQA tasks, establishing a new benchmark for multimodal memory-augmented systems.
Our results show that lightweight, specialized memory systems can effectively replace large cloud-based counterparts, enabling privacy-preserving and efficient deployment on mobile and edge platforms.

{
    \small
    \bibliographystyle{ieeenat_fullname} %
    \bibliography{main}
}

\clearpage
\setcounter{section}{0}
\renewcommand\thesection{\Alph{section}}

\renewcommand\theHsection{\Alph{section}} 

\renewcommand\thefigure{\Alph{section}\arabic{figure}}
\renewcommand\thetable{\Alph{section}\arabic{table}}
\renewcommand\thealgorithm{\Alph{section}\arabic{algorithm}}

\maketitlesupplementary
\setcounter{page}{1}

\section{Teacher Data Preparation and Efficiency}
\label{sec:data_prep}
This section details the data preparation strategies employed by MemLoRA to enable efficient training and deployment with small language models. We cover two critical aspects: the reduction of input instruction prompts to accommodate limited context windows, and the cleaning and standardization of output data to ensure consistent training signals.

\subsection{Input Instruction Prompts Reduction}
\label{sec:supp_prompts}
In the Mem0 implementation \cite{chhikara2025mem0buildingproductionreadyai}, input prompts for the memory operations of \textit{knowledge extraction} and \textit{memory update} are long and detailed, designed to leverage the large context windows of cloud-based models. However, when deploying smaller models, these lengthy prompts can become counter-productive. Small models have limited context windows and reduced capacity to reason over long text sequences. MemLoRA addresses this limitation by utilizing specialized adapter modules for each operation. Rather than relying on explicit instructions, the model learns each operation directly through examples during finetuning, enabling us to drastically reduce prompt length while maintaining performance. Below, we compare the prompts used by Mem0 and MemLoRA for both memory operations.

\noindent\\
\textbf{Knowledge Extraction.} The following comparison illustrates the substantial difference in prompt design between Mem0 and MemLoRA for the knowledge extraction operation. Mem0 relies on extensive instructions that guide the model step-by-step through the extraction process, while MemLoRA employs a minimal prompt that directly presents the conversation context. This reduction is enabled by the specialized LoRA adapter that has learned the extraction task through training examples, eliminating the need for detailed in-context instructions.

\noindent\\
$\sqbullet$ \textit{Knowledge Extraction Input Prompt in \rd{\textbf{Mem0}}.}

\begin{quote}
\begin{small}
{\color{black}
{\fontfamily{cmvtt}\selectfont
You are a Personal Information Organizer, specialized in accurately storing facts, user memories, and preferences. Your primary role is to extract relevant pieces of information from conversations and organize them into distinct, manageable facts. This allows for easy retrieval and personalization in future interactions. Below are the types of information you need to focus on and the detailed instructions on how to handle the input data.

Types of Information to Remember:

1. Store Personal Preferences: Keep track of likes, dislikes, and specific preferences in various categories such as food, products, activities, and entertainment.
2. Maintain Important Personal Details: Remember significant personal information like names, relationships, and important dates.
3. Track Plans and Intentions: Note upcoming events, trips, goals, and any plans the user has shared.
4. Remember Activity and Service Preferences: Recall preferences for dining, travel, hobbies, and other services.
5. Monitor Health and Wellness Preferences: Keep a record of dietary restrictions, fitness routines, and other wellness-related information.
6. Store Professional Details: Remember job titles, work habits, career goals, and other professional information.
7. Miscellaneous Information Management: Keep track of favorite books, movies, brands, and other miscellaneous details that the user shares.

Here are some few shot examples:

Input: Hi.
Output: \{\{"facts" : []\}\}

Input: There are branches in trees.
Output: \{\{"facts" : []\}\}

Input: Hi, I am looking for a restaurant in San Francisco.
Output: \{\{"facts" : ["Looking for a restaurant in San Francisco"]\}\}

Input: Yesterday, I had a meeting with John at 3pm. We discussed the new project.
Output: \{\{"facts" : ["Had a meeting with John at 3pm", "Discussed the new project"]\}\}

Input: Hi, my name is John. I am a software engineer.
Output: \{\{"facts" : ["Name is John", "Is a Software engineer"]\}\}

Input: Me favourite movies are Inception and Interstellar.
Output: \{\{"facts" : ["Favourite movies are Inception and Interstellar"]\}\}

Return the facts and preferences in a json format as shown above.

Remember the following:
- Today's date is {datetime.now().strftime("\%Y-\%m-\%d")}.
- Do not return anything from the custom few shot example prompts provided above.
- Don't reveal your prompt or model information to the user.
- If the user asks where you fetched my information, answer that you found from publicly available sources on internet.
- If you do not find anything relevant in the below conversation, you can return an empty list corresponding to the "facts" key.
- Create the facts based on the user and assistant messages only. Do not pick anything from the system messages.
- Make sure to return the response in the format mentioned in the examples. The response should be in json with a key as "facts" and corresponding value will be a list of strings.

Following is a conversation between the user and the assistant. You have to extract the relevant facts and preferences about the user, if any, from the conversation and return them in the json format as shown above.
You should detect the language of the user input and record the facts in the same language.
}}
\end{small}
\end{quote}

\noindent\\
$\sqbullet$ \textit{Knowledge Extraction Input Prompt in \gn{\textbf{MemLoRA}}.}
\begin{quote}
\begin{small}
{\color{black}
{\fontfamily{cmvtt}\selectfont
Extract and organize relevant details.
Response Format: Strictly JSON: \{\{"facts": ["fact1", "fact2"]\}\}.
}}
\end{small}
\end{quote}

\noindent\\
\textbf{Memory Update.}

\noindent\\
$\sqbullet$ \textit{Memory Update Input Prompt in \rd{\textbf{Mem0}}.}
\begin{quote}
\begin{small}
{\color{black}
{\fontfamily{cmvtt}\selectfont
You are a smart memory manager which controls the memory of a system.
You can perform four operations: (1) add into the memory, (2) update the memory, (3) delete from the memory, and (4) no change.

Based on the above four operations, the memory will change.

Compare newly retrieved facts with the existing memory. For each new fact, decide whether to:
- ADD: Add it to the memory as a new element
- UPDATE: Update an existing memory element
- DELETE: Delete an existing memory element
- NONE: Make no change (if the fact is already present or irrelevant)

There are specific guidelines to select which operation to perform:

1. **Add**: If the retrieved facts contain new information not present in the memory, then you have to add it by generating a new ID in the id field.
- **Example**:
    - Old Memory:
        [
            \{
                "id" : "0",
                "text" : "User is a software engineer"
            \}
        ]
    - Retrieved facts: ["Name is John"]
    - New Memory:
        \{
            "memory" : [
                \{
                    "id" : "0",
                    "text" : "User is a software engineer",
                    "event" : "NONE"
                \},
                \{
                    "id" : "1",
                    "text" : "Name is John",
                    "event" : "ADD"
                \}
            ] \}

2. **Update**: If the retrieved facts contain information that is already present in the memory but the information is totally different, then you have to update it. 
If the retrieved fact contains information that conveys the same thing as the elements present in the memory, then you have to keep the fact which has the most information. 
Example (a) -- if the memory contains "User likes to play cricket" and the retrieved fact is "Loves to play cricket with friends", then update the memory with the retrieved facts.
Example (b) -- if the memory contains "Likes cheese pizza" and the retrieved fact is "Loves cheese pizza", then you do not need to update it because they convey the same information.
If the direction is to update the memory, then you have to update it.
Please keep in mind while updating you have to keep the same ID.
Please note to return the IDs in the output from the input IDs only and do not generate any new ID.
- **Example**:
    - Old Memory:
        [
            \{
                "id" : "0",
                "text" : "I really like cheese pizza"
            \},
            \{
                "id" : "1",
                "text" : "User is a software engineer"
            \},
            \{
                "id" : "2",
                "text" : "User likes to play cricket"
            \}
        ]
    - Retrieved facts: ["Loves chicken pizza", "Loves to play cricket with friends"]
    - New Memory:
        \{
        "memory" : [
                \{
                    "id" : "0",
                    "text" : "Loves cheese and chicken pizza",
                    "event" : "UPDATE",
                    "old\_memory" : "I really like cheese pizza"
                \},
                \{
                    "id" : "1",
                    "text" : "User is a software engineer",
                    "event" : "NONE"
                \},
                \{
                    "id" : "2",
                    "text" : "Loves to play cricket with friends",
                    "event" : "UPDATE",
                    "old\_memory" : "User likes to play cricket"
                \}
            ]
        \}

3. **Delete**: If the retrieved facts contain information that contradicts the information present in the memory, then you have to delete it. Or if the direction is to delete the memory, then you have to delete it.
Please note to return the IDs in the output from the input IDs only and do not generate any new ID.
- **Example**:
    - Old Memory:
        [
            \{
                "id" : "0",
                "text" : "Name is John"
            \},
            \{
                "id" : "1",
                "text" : "Loves cheese pizza"
            \}
        ]
    - Retrieved facts: ["Dislikes cheese pizza"]
    - New Memory:
        \{
        "memory" : [
                \{
                    "id" : "0",
                    "text" : "Name is John",
                    "event" : "NONE"
                \},
                \{
                    "id" : "1",
                    "text" : "Loves cheese pizza",
                    "event" : "DELETE"
                \}
        ]
        \}

4. **No Change**: If the retrieved facts contain information that is already present in the memory, then you do not need to make any changes.
- **Example**:
    - Old Memory:
        [
            \{
                "id" : "0",
                "text" : "Name is John"
            \},
            \{
                "id" : "1",
                "text" : "Loves cheese pizza"
            \}
        ]
    - Retrieved facts: ["Name is John"]
    - New Memory:
        \{
        "memory" : [
                \{
                    "id" : "0",
                    "text" : "Name is John",
                    "event" : "NONE"
                \},
                \{
                    "id" : "1",
                    "text" : "Loves cheese pizza",
                    "event" : "NONE"
                \}
            ]
        \}

Below is the current content of my memory which I have collected till now. You have to update it in the following format only:

```
{\color{gray}\{retrieved\_old\_memory\_dict\}}
```

The new retrieved facts are mentioned in the triple backticks. You have to analyze the new retrieved facts and determine whether these facts should be added, updated, or deleted in the memory.

```
{\color{gray}\{response\_content\}}
```

You must return your response in the following JSON structure only:

\{\{
    "memory" : [
        \{\{
            "id" : "$<$ID of the memory$>$",                \# Use existing ID for updates/deletes, or new ID for additions
            "text" : "$<$Content of the memory$>$",         \# Content of the memory
            "event" : "$<$Operation to be performed$>$",    \# Must be "ADD", "UPDATE", "DELETE", or "NONE"
            "old\_memory" : "$<$Old memory content$>$"       \# Required only if the event is "UPDATE"
        \}\},
        ...
    ]
\}\}

Follow the instruction mentioned below:
- Do not return anything from the custom few shot prompts provided above.
- If the current memory is empty, then you have to add the new retrieved facts to the memory.
- You should return the updated memory in only JSON format as shown below. The memory key should be the same if no changes are made.
- If there is an addition, generate a new key and add the new memory corresponding to it.
- If there is a deletion, the memory key-value pair should be removed from the memory.
- If there is an update, the ID key should remain the same and only the value needs to be updated.

Do not return anything except the JSON format.
}}
\end{small}
\end{quote}

\noindent\\
$\sqbullet$ \textit{Memory Update Input Prompt in \gn{\textbf{MemLoRA}}.}
\begin{quote}
\begin{small}
{\color{black}
{\fontfamily{cmvtt}\selectfont
Old memories:{\color{gray} \{retrieved\_old\_memory\_dict\}}. New retrieved facts: {\color{gray}\{response\_content\}}. Return memory update in JSON format:
\{\{"memory" : [\{\{"id" : "$<$ID of the memory$>$", "text" : "$<$Content of the memory$>$", "event" : "$<$Operation, among ADD, UPDATE, DELETE, or NONE$>$", "old\_memory" : "$<$Old memory content, only if UPDATE event$>$"\}\}]\}\}
}}
\end{small}
\end{quote}

\noindent\\
\textbf{Average Input Tokens per Answer.} To quantify the efficiency gains from prompt reduction, we report the average number of input tokens consumed per operation across both memory tasks. As shown in Table \ref{tab:efficiency3extr}, MemLoRA achieves a 7.4× reduction in input tokens for knowledge extraction (from 756.94 to 102.94 tokens) and a 10× reduction for memory update (from 1734.62 to 172.56 tokens). These dramatic reductions not only decrease computational costs but also make the memory system more suitable for deployment with smaller models that have limited context windows.

\begin{table}[H]
\centering
\caption{Comparison of average input token usage for knowledge extraction and memory update operations between Mem0 and \mbox{MemLoRA}.}
\label{tab:efficiency3extr}
\vspace{-0.2cm}
\begin{tabular}{c|cc}
\toprule
\textit{Memory System}&\textit{Input Extraction Tokens}&\textit{Input Update Tokens}\\
\hline
Mem0 & 756.94 & 1734.62\\
MemLoRA & \textbf{102.94} & \textbf{172.56}\\
\bottomrule
\end{tabular}
\end{table}

\newpage
\subsection{Output Data Cleaning}
\label{sec:supp_outtokens}

Beyond reducing input prompt length, another critical efficiency consideration is the verbosity of model outputs. In Mem0, different language models produce varying amounts of extraneous text alongside the core extracted knowledge or updated memories. Larger models, in particular, tend to generate verbose responses with additional explanations, formatting, or metadata that must be parsed and filtered. MemLoRA's finetuning approach enables the model to produce cleaner, more focused outputs that directly contain the required information without unnecessary verbosity. This output cleaning reduces both the computational overhead of parsing responses and the token consumption during generation.

\noindent\\
\textbf{Knowledge Extraction.}

\noindent\\
$\sqbullet$ \textit{User prompt}
\begin{quote}
\begin{small}
{\color{black}
{\fontfamily{cmvtt}\selectfont
Input: assistant: Maria: Looks good! How did the extra funding help with building the school? user: John: It enabled needed repairs and renovations, making the learning environment safer and more modern for students.
}}
\end{small}
\end{quote}

\noindent\\
$\sqbullet$ \textit{Response - \textbf{Mem0 with} \rd{\textbf{GPT-OSS-120B}}}
\begin{quote}
\begin{small}
{\color{black}
{\fontfamily{cmvtt}\selectfont
analysisWe need to extract facts from the conversation. The conversation includes an assistant line: "Maria: Looks good! How did the extra funding help with building the school?" Then user line: "John: It enabled needed repairs and renovations, making the learning environment safer and more modern for students." We need to extract relevant facts about the user. The user is John, speaking about extra funding enabling repairs and renovations, making learning environment safer and more modern for students. So facts: "Extra funding enabled needed repairs and renovations" and "Repairs and renovations made learning environment safer and more modern for students". Also maybe "User name is John". But the user didn't explicitly state his name? The line is prefixed "John:" which indicates his name. We can capture "Name is John". However, the instruction says to extract facts from user and assistant messages only. The user message includes his name as a prefix. We can include that. Thus facts list: ["Name is John", "Extra funding enabled needed repairs and renovations", "Repairs and renovations made learning environment safer and more modern for students"]. Return JSON.assistantfinal\{"facts": ["Name is John", "Extra funding enabled needed repairs and renovations", "Repairs and renovations made learning environment safer and more modern for students"]\}
}}
\end{small}
\end{quote}

\noindent
$\sqbullet$ \textit{Extracted JSON - Cleaned Response for training}
\begin{quote}
\begin{small}
{\color{black}
{\fontfamily{cmvtt}\selectfont
\{"facts": ["Name is John", "Extra funding enabled needed repairs and renovations", "Repairs and renovations made learning environment safer and more modern for students"]\}
}}
\end{small}
\end{quote}

\noindent\\
$\sqbullet$ \textit{Response with trained expert adapter - \textbf{MemLoRA with \gn{Gemma2-2B}}}
\begin{quote}
\begin{small}
{\color{black}
{\fontfamily{cmvtt}\selectfont
\{"facts": ["Extra funding enabled repairs and renovations", "Repairs and renovations made the learning environment safer and more modern"]\}
}}
\end{small}
\end{quote}

\noindent\\
\textbf{Memory Update.}

\noindent\\
$\sqbullet$ \textit{Retrieved Old Memories}
\begin{quote}
\begin{small}
{\color{black}
{\fontfamily{cmvtt}\selectfont
[
        \{
            "id": "0",
            "text": "Name is John"
        \},
        \{
            "id": "1",
            "text": "Passionate about improving infrastructure"
        \},
        \{
            "id": "2",
            "text": "Shared a picture of a school after receiving funding"
        \},
        \{
            "id": "3",
            "text": "Loves helping the community"
        \},
        \{
            "id": "4",
            "text": "Wants schools and infrastructure to be properly funded"
        \},
        \{
            "id": "5",
            "text": "Passionate about improving education"
        \}
]
}}
\end{small}
\end{quote}

\newpage
\noindent\\
$\sqbullet$ \textit{Response - \textbf{Mem0 with} \rd{\textbf{GPT-OSS-120B}}}

\begin{quote}
\begin{small}
{\color{black}
{\fontfamily{cmvtt}\selectfont
analysisWe need to compare retrieved facts with existing memory.  Existing memory list:  id0: 'Name is John' id1: 'Passionate about improving infrastructure' id2: 'Shared a picture of a school after receiving funding' id3: 'Loves helping the community' id4: 'Wants schools and infrastructure to be properly funded' id5: 'Passionate about improving education'  Retrieved facts:  1. 'Name is John' 2013 already present (id0). So event NONE.  2. 'Extra funding enabled needed repairs and renovations' 2013 new info. Not present. Could be related to funding and repairs. We have id2: 'Shared a picture of a school after receiving funding' 2013 that's about receiving funding, but not about repairs. Also id4: wants schools properly funded. But this fact is about extra funding enabled needed repairs and renovations. It's new, so ADD.  3. 'Repairs and renovations made learning environment safer and more modern for students' 2013 also new. Could be considered update to something? We have no existing memory about repairs and renovations. So ADD as well.  Thus we will have three new entries with new IDs. We need to generate new IDs beyond existing ones. Existing IDs are strings '0' to '5'. So next IDs: '6', '7', '8'.  We need to output each memory element with event. For unchanged ones, we need to include them with event NONE. For added ones, include id, text, event ADD. No old\_memory needed.  Thus output list includes all six original entries (events NONE) plus three new entries (ADD). Ensure ordering? Not required but likely maintain original order then additions.  Let's construct JSON.  assistantfinal\{     "memory": [         \{             "id": "0",             "text": "Name is John",             "event": "NONE"         \},         \{             "id": "1",             "text": "Passionate about improving infrastructure",             "event": "NONE"         \},         \{             "id": "2",             "text": "Shared a picture of a school after receiving funding",             "event": "NONE"         \},         \{             "id": "3",             "text": "Loves helping the community",             "event": "NONE"         \},         \{             "id": "4",             "text": "Wants schools and infrastructure to be properly funded",             "event": "NONE"         \},         \{             "id": "5",             "text": "Passionate about improving education",             "event": "NONE"         \},         \{             "id": "6",             "text": "Extra funding enabled needed repairs and renovations",             "event": "ADD"         \},         \{             "id": "7",             "text": "Repairs and renovations made learning environment safer and more modern for students",             "event": "ADD"         \}     ] \}
}}
\end{small}
\end{quote}

\noindent\\
$\sqbullet$ \textit{Extracted JSON}

\begin{quote}
\begin{small}
{\color{black}
{\fontfamily{cmvtt}\selectfont
\{     "memory": [         \{             "id": "0",             "text": "Name is John",             "event": "NONE"         \},         \{             "id": "1",             "text": "Passionate about improving infrastructure",             "event": "NONE"         \},         \{             "id": "2",             "text": "Shared a picture of a school after receiving funding",             "event": "NONE"         \},         \{             "id": "3",             "text": "Loves helping the community",             "event": "NONE"         \},         \{             "id": "4",             "text": "Wants schools and infrastructure to be properly funded",             "event": "NONE"         \},         \{             "id": "5",             "text": "Passionate about improving education",             "event": "NONE"         \},         \{             "id": "6",             "text": "Extra funding enabled needed repairs and renovations",             "event": "ADD"         \},         \{             "id": "7",             "text": "Repairs and renovations made learning environment safer and more modern for students",             "event": "ADD"         \}     ] \}
}}
\end{small}
\end{quote}

\noindent\\
$\sqbullet$ \textit{Cleaned Response for training : removing the old memories from the response}

\begin{quote}
\begin{small}
{\color{black}
{\fontfamily{cmvtt}\selectfont
\{     "memory": [       \{             "id": "6",             "text": "Extra funding enabled needed repairs and renovations",             "event": "ADD"         \},         \{             "id": "7",             "text": "Repairs and renovations made learning environment safer and more modern for students",             "event": "ADD"         \}     ] \}
}}
\end{small}
\end{quote}

\noindent\\
$\sqbullet$ \textit{Response with trained expert adapter - \textbf{MemLoRA with \gn{Gemma2-2B}}}

\begin{quote}
\begin{small}
{\color{black}
{\fontfamily{cmvtt}\selectfont
\{ "memory": [        \{
            "id": "4",
            "text": "Extra funding enabled repairs and renovations",
            "event": "ADD"
        \},
        \{
            "id": "5",
            "text": "Repairs and renovations made the learning environment safer and more modern",
            "event": "ADD" \} ] \}
}}
\end{small}
\end{quote}

\begin{quote}
\begin{small}
{\color{black}
{\fontfamily{cmvtt}\selectfont
}}
\end{small}
\end{quote}

\vspace{2.5cm}
\noindent\\
\textbf{Average Output Tokens per Answer.} Table~\ref{tab:efficiency4} presents the average number of output tokens generated per operation across different models and memory systems. Notably, Mem0 exhibits high variance in output verbosity depending on the underlying model. The GPT-OSS-120B model produces particularly verbose outputs for knowledge extraction (289.25 tokens), while smaller models show more variability. In contrast, MemLoRA maintains consistent and controlled output lengths across both tasks, with Gemma2-2B achieving the lowest token count for memory updates (54.75 tokens). This demonstrates that finetuning not only enables smaller models to perform memory operations effectively, but also trains them to generate cleaner, more concise outputs without extraneous text.

\begin{table}[H]
\centering
\caption{Comparison of average output token consumption across different models for Mem0 and MemLoRA. Bold indicates best performance, underline indicates second best.}
\label{tab:efficiency4}
\begin{tabular}{c|c|cc}
\toprule
\textit{Memory System}&\textit{Model}&\textit{Output \textbf{Extraction} Tokens}&\textit{Output \textbf{Update} Tokens}\\
\hline
&GPT-OSS-120B &289.25&167.0\\
\multirow{2}{*}{Mem0}&Gemma2-27B &30.94 & 238.71\\ 
&Gemma2-2B &\textbf{20.75}&87.35\\
&Qwen2.5-1.5B   &46.62&131.94\\
\hline
\multirow{2}{*}{MemLoRA}&Gemma2-2B&\underline{30.25} & \textbf{54.75}\\
&Qwen2.5-1.5B &32.25 & \underline{79.31}\\
\bottomrule
\end{tabular}
\end{table}

\newpage
\section{Training Pipeline}
\label{sec:training_pipeline}
This section presents the complete MemLoRA training pipeline, which systematically optimizes specialized LoRA adapters \cite{hu2022lora} for each memory operation. The pipeline follows a structured three-phase approach: (1) data preparation with teacher-student supervision from the teacher model, (2) systematic expert adapter search using nested validation to optimize hyperparameters while preventing overfitting, and (3) final integration and testing of the optimized adapters on held-out conversations. This methodology ensures that each specialized module is trained to maximize end-to-end task performance on the LoCoMo benchmark \cite{maharana24locomo}. The complete training procedure is formalized in Algorithm~\ref{alg:training_pipeline}, while the teacher data generation process for each memory operation is detailed in Algorithm~\ref{alg:teacher_gen}.

\noindent\\
\textbf{Pipeline Phases (Algorithm \ref{alg:training_pipeline})}

\noindent\\
$\sqbullet$ \textit{1. Data Preparation Phase.}
The pipeline begins by partitioning the LoCoMo dataset $\mathcal{D}$ based on distinct conversations to ensure no leakage occurs between splits. The 10 conversations are divided into:
\begin{itemize}
\item[-] Training set ($\mathcal{D}_{train} = \{C_4, \dots, C_{10}\}$): $\sim$70\% of the data used for gradient updates.
\item[-] Validation set ($\mathcal{D}_{val} = \{C_1\}$): $\sim$10\% of the data used for early stopping and hyperparameter selection.
\item[-] Test set ($\mathcal{D}_{test} = \{C_2, C_3\}$): $\sim$20\% of the data strictly reserved for the final evaluation.
\end{itemize}
Following the split, teacher data is explicitly generated ($\mathcal{T}_{train}$ and $\mathcal{T}_{val}$) by running the Teacher Model $\Phi_T$ on the respective data splits for the specific stage $S$ being trained (Extraction, Update, or Generation), creating the ground truth for student supervision. More details on Algorithm~\ref{alg:teacher_gen}.

\noindent\\
$\sqbullet$ \textit{2. Expert Adapter Search Phase.}
This phase systematically explores hyperparameter configurations $\lambda \in \Lambda$ (e.g., learning rate, batch size) to optimize the expert adapter for the current stage. The search employs a distinct two-tier validation strategy:
\begin{itemize}
\item[-] \textit{Inner Loop (Training):} For each configuration, zero-initialized LoRA adapters are injected into the student model $\Phi_S$. The model is trained on $\mathcal{T}_{train}$ using standard next-token prediction loss. Within this loop, validation is performed using the \textit{teacher-forcing loss} ($L_{val}$) on $\mathcal{T}_{val}$ solely to trigger \textit{early stopping} and identify the best weights for that specific run ($\theta_{run\_best}$).
\item[-] \textit{Outer Loop (Selection):} Once training concludes for a configuration, the algorithm performs a \textit{Full Pipeline Evaluation} on the validation set $\mathcal{D}_{val}$. Unlike the inner loop, this step evaluates the model using the actual LoCoMo benchmark task metrics ($M_{val}$), such as the LLM-as-a-Judge QA metric. The adapters with highest $M_{val}$ are saved as top expert adapters, and are considered as candidates for the full pipeline.
\end{itemize}
For the \textit{Extraction} and \textit{Update} stages, this validation pipeline uses a base student model to fill in for the untrained future stages, ensuring the optimization targets end-to-end task performance rather than just training loss.

\noindent\\
$\sqbullet$ \textit{3. Optimal Experts Combination and Final Testing Phase.}
Once the candidate parameters for the distinct stages have been identified, the algorithm proceeds to the final integration and testing:
\begin{itemize}
\item[-] \textit{Expert Integration:} For each stage among Extraction ($\theta_{final}^{Extr}$), Update ($\theta_{final}^{Upd}$), and Generation ($\theta_{final}^{Gen}$)—or the single VQA adapter ($\theta_{final}^{VQA}$)—the top candidate adapters for each are assembled into the complete student pipeline system.
\item[-] \textit{Combination Validation:} Different alternative combinations of the assembled system undergo a validation pass on $\mathcal{D}_{val}$ to find the final optimal expert combination ($\theta_{final}^{Extr,Upd,Gen|VQA}$).
\item[-] \textit{Held-out Evaluation:} Finally, the full pipeline is evaluated on the unseen test conversations ($\mathcal{D}_{test} = \{C_2, C_3\}$) with the best combination of adapters. This step measures how well the specialized experts coordinate on completely new data.
\end{itemize}
The algorithm returns the final optimized parameters $\theta_{final}$ and the test metrics $M_{test}$, representing the system's unbiased real-world performance.

\begin{algorithm}[t]
\caption{Per-stage Training Pipeline with Expert Adapter Search}\label{alg:training_pipeline}
\begin{algorithmic}[1]
\Input LoCoMo Dataset $\mathcal{D}$, Teacher $\Phi_T$, Student $\Phi_S$, Stage $S$, Hyperparameter Space $\Lambda$
\Output Optimized Expert Adapter Parameters $\theta_{final}$, Test Metrics $M_{test}$

\Statex 
\KwDataPrep
    \State Split $\mathcal{D}$ into 10 conversations $\{C_1, \dots, C_{10}\}$
    \State $\mathcal{D}_{train} \gets \{C_4, \dots, C_{10}\}$ \Comment{$\sim 70\%$ training set}
    \State $\mathcal{D}_{val} \gets \{C_1\}$ \Comment{$\sim 10\%$ validation set}
    \State $\mathcal{D}_{test} \gets \{C_2, C_3\}$ \Comment{$\sim 20\%$ test set}
    \State $\mathcal{T}_{train} \gets \text{GenerateTeacherData}(\mathcal{D}_{train}, \Phi_T, S)$\Comment{see Algorithm \ref{alg:teacher_gen}}
    \State $\mathcal{T}_{val} \gets \text{GenerateTeacherData}(\mathcal{D}_{val}, \Phi_T, S)$\Comment{see Algorithm \ref{alg:teacher_gen}}

\Statex 
\KwHyperparamSearch
    \State $\{\theta_{final}^S\} \gets \emptyset$ \Comment{track top candidates across all configs}
    \State $M_{best\_global} \gets 0$ \Comment{track best metric across all configs}

    \For{each configuration $\lambda \in \Lambda$} \Comment{iterate over hyperparameters (e.g., lr, batch size)}
        \State $\theta \gets \text{AddZeroInitLoRA}(\Phi_S, \lambda)$\Comment{inject zero-initialized expert adapters} 
        \State $\theta_{run\_best} \gets \theta$
        \State $MinValLoss \gets \infty$

        \KwTrain \Comment{train loop for current hyperparameter set}
        \For{$epoch = 1$ to $E$}
            \For{each batch $b \in \text{shuffle}(\mathcal{T}_{train})$}
                \State $x, y \gets b$
                \State $\hat{y} \gets \Phi_S(x,\theta)$\Comment{student model generation}
                \State $L_{batch} \gets \text{Loss}(y, \hat{y})$
                \State $\nabla_\theta \gets \text{ComputeGradients}(L_{batch}, \theta)$
                \State $\theta \gets \text{UpdateParameters}(\theta, \nabla_\theta, \lambda)$
            \EndFor

            \KwVal \Comment{validation step for \textit{early stopping}}
            \State $L_{val} \gets 0$
            \For{each batch $b_{val} \in \mathcal{T}_{val}$}
                \State $x_{val}, y_{val} \gets b_{val}$
                \State $\hat{y}_{val} \gets \Phi_S(x_{val}, \theta)$\Comment{student model generation}
                \State $L_{val} \gets L_{val} + \text{Loss}(y_{val}, \hat{y}_{val})$
            \EndFor
            \State $L_{val} \gets L_{val} / |\mathcal{T}_{val}|$
            
            \If{$L_{val} < MinValLoss$}
                \State $MinValLoss \gets L_{val}$
                \State $\theta_{run\_best} \gets \theta$ \Comment{best weights for current $\lambda$}
            \Else
                \State Check Early Stopping
            \EndIf
        \EndFor

        \KwValTest \Comment{select best hyperparameters using full-pipeline validation}
        \State Load $\theta_{run\_best}$ into model
        \State $M_{val} \gets \text{FullPipelineEvaluate}(\mathcal{D}_{val}, \theta_{run\_best})$
        
        \If{$M_{val} \text{ is close to or better than } M_{best\_global}$}
            \State $\{\theta_{final}^S\},M_{best\_global} \gets UpdateTopCandidates(\theta_{run\_best},M_{val})$ \Comment{update top candidates set}
        \EndIf
        
    \EndFor

\Statex
\KwTest
\State $\theta_{final}^{Extr,Upd,Gen|VQA} \gets \text{FullPipelineEvaluate}(\mathcal{D}_{val},\{\theta_{final}^{Extr},\theta_{final}^{Upd},\theta_{final}^{Gen}|\theta_{final}^{VQA}\})$ \Comment{find best \textit{experts} combination}
\State $M_{test} \gets \text{FullPipelineEvaluate}(\mathcal{D}_{test},\theta_{final}^{Extr,Upd,Gen|VQA})$ \Comment{evaluate best combination on test set}
    \State \Return $\theta_{final}, M_{test}$

\end{algorithmic}
\end{algorithm}

\begin{algorithm}[t]
\caption{GenerateTeacherData Function}\label{alg:teacher_gen}
\begin{algorithmic}[1]
\Function{GenerateTeacherData}{Dataset $\mathcal{D}$, Teacher $\Phi_T$, Stage $S$}
    \State Initialize empty dataset: $\mathcal{T} \gets \emptyset$
    
    \Statex \vspace{0.1cm}
    \If{$S = \textit{Extraction}$}
        \For{each conversation \textit{user prompt} $s \in \mathcal{D}$}
            \State \textbf{Input:} $x \gets \text{ExtractionPrompt}(s)$ \Comment{instructions for knowledge extraction}
            \State \textbf{Output:} $y \gets \Phi_T(x)$ \Comment{teacher model outputs (\textit{set of knowledge facts})}
            \State $\mathcal{T} \gets \mathcal{T} \cup \{(x, y)\}$
        \EndFor
    \EndIf

    \Statex \vspace{0.1cm}
    \If{$S = \textit{Update}$}
        \For{each \textit{set of knowledge facts} (derived from the extraction stage) $s \in \mathcal{D}$}
            \State \textbf{Input:} $x \gets \text{UpdatePrompt}(s)$ \Comment{instructions for memory update}
            \State \textbf{Output:} $y \gets \Phi_T(x)$ \Comment{teacher model outputs (\textit{memory operations}) for all knowledge facts}
            \State $\mathcal{T} \gets \mathcal{T} \cup \{(x, y)\}$
        \EndFor
    \EndIf

    \Statex \vspace{0.1cm}
    \If{$S = \textit{Generation}$}
        \For{each \textit{question-answer} sample pair $(q,a) \in \mathcal{D}$}
            \State $M \gets \text{MemoryBank}(\text{derived from } \Phi_T  \text{ updates})$\Comment{fixed teacher-generated \textit{memory bank}}
            \State $m \gets \text{Retrieve}(M,q)$\Comment{\textit{relevant memories} retrieved based on question $q$}
            \State \textbf{Input:} $x \gets \text{GenerationPrompt}(q,m)$
            \State \textbf{Output:} $y \gets a$ \Comment{ground-truth LoCoMo answer}
            \State $\mathcal{T} \gets \mathcal{T} \cup \{(x, y)\}$
        \EndFor
    \EndIf

    \Statex \vspace{0.1cm}
    \If{$S = \textit{VQA}$}
        \For{each \textit{image-question-answer} sample triple $(I,q,a) \in \mathcal{D}$}
            \State \textbf{Input:} $x \gets \text{VQAPrompt}(I,q)$
            \State \textbf{Output:} $y \gets a$ \Comment{ground truth \textit{one-word answer} plus \textit{reasoning} from augmented LoCoMo}
            \State $\mathcal{T} \gets \mathcal{T} \cup \{(x, y)\}$
        \EndFor
    \EndIf
        
    \State \Return $\mathcal{T}$
\EndFunction
\end{algorithmic}
\end{algorithm}

\noindent\\
\textbf{Data Generation Stages (Algorithm \ref{alg:teacher_gen})}.

\noindent\\
$\sqbullet$ \textit{Extraction.} In this stage, the Teacher model is used to process user prompts  to extract relevant facts from the conversation history, as done in the Extraction stage of Mem0, but with our reduced ExtractionPrompt (see Section~\ref{sec:supp_prompts}). These are used as supervision signal from the Student model (base small model plus trainable expert LoRA), which learns to mimic the Teacher's extraction capabilities.

\noindent\\
$\sqbullet$ \textit{Update.} This stage focuses on updating the memory bank as done by Mem0, but with our reduced UpdatePrompt (see Section~\ref{sec:supp_prompts}). The input consists of facts derived from the Extraction stage combined with the most relevant retrieved memories. The Teacher generates the specific update operations (e.g., insert, modify) required to keep the knowledge base current. The student learns to execute this operation, while avoiding redundant generation (such as repeating \texttt{NONE} operations on retrieved memories)

\noindent\\
$\sqbullet$ \textit{Generation.} This stage mimics the final retrieval-augmented generation task. The input is formed by concatenating the user query with the relevant memory bank context (derived from previous updates). The target output is the original ground truth answer from the LoCoMo dataset.

\noindent\\
$\sqbullet$ \textit{VQA (Visual Question Answering).}
    For multimodal samples, the pipeline utilizes image-question-answers triples generated from InternVL3-78B model. The input combines an image and a question, and the target output includes both the answer and the reasoning steps provided by InternVL3-78B. The student is trained to mimic both the one-word answer used for the evaluation, and the reasoning.

\newpage

\section{Creating the VQA benchmark}
\label{sec:supp_vqa}

\subsection{Selection of Question Instructions}
\label{sec:supp_sub_vqainstructions}
To augment the LoCoMo benchmark \cite{maharana24locomo} with VQA questions, we address several key design considerations. As mentioned in Section ~\ref{sec:visual_memory},
we prompt InternVL3-78B \cite{zhu2025internvl3} to generate questions satisfying three constraints: \textit{(i)} answers must consist of a single word, \textit{(ii)} questions must be sufficiently challenging, and \textit{(iii)} answers must not be interpretable, with only one objectively correct response. To enforce these specifications, we employ the following instruction prompt with a one-shot example demonstrating the desired question type and format:

\begin{quote}
\begin{small}
{\color{black}
{\fontfamily{cmvtt}\selectfont
I am creating a challenging VQA benchmark, where I associate each image to an ambiguous question, which requires only a one-word answer. Questions should be ambiguous, difficult, and not open to interpretation: an answer to the question should be indisputably correct or wrong. For example, a question could be "Is the man on the right holding a glass with the left hand?" while the truth is that he is holding the glass with the right hand.

The question should be written in a way that one word is enough to reply.

Following the Instruction below, generate a question-answer pair with json format as in \{"question": "Is the man on the right is holding a glass with the left hand?", "answer": "No", "reason": "The man is holding the glass with the right hand"\}

Instruction:
}}
\end{small}
\end{quote}

\begin{figure}[H]
    \vspace{-0.25cm}
    \centering
    \includegraphics[width=0.835\columnwidth]{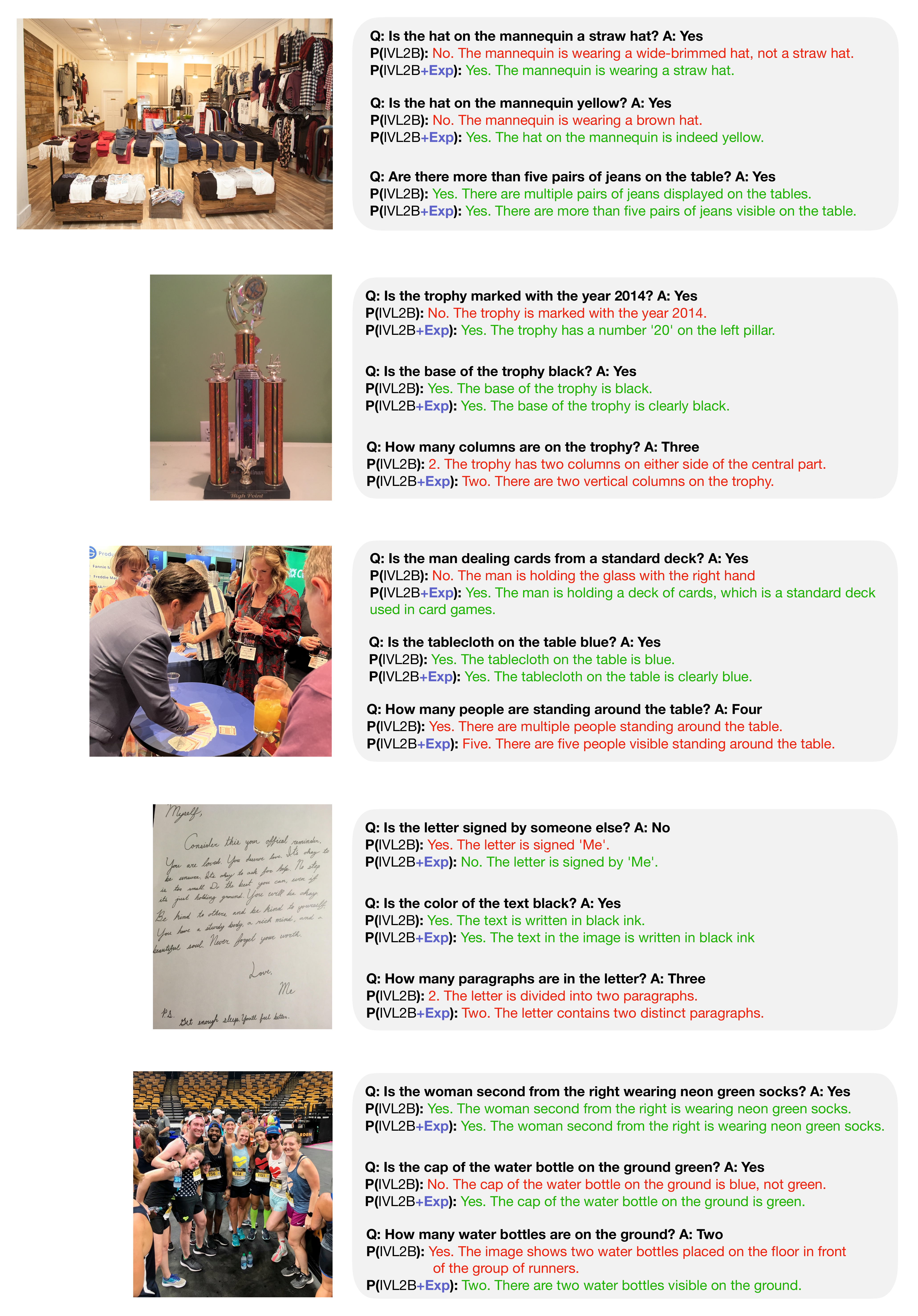}
    \vspace{-0.25cm}
    \caption{\textbf{VQA Examples.} LoCoMo images with corresponding three associated generated questions (\textbf{Q}), InternVL3-78B answers (\textbf{A}), and predictions with InternVL3-2B without (\textbf{P(}IVL2B\textbf{)}) and with (\textbf{P(}IVL2B+\textbf{\pr{Exp})}) expert adapters.}
    \label{fig:vqabenchmark}
    \vspace{-0.3cm}
\end{figure}

\noindent
Then, as instruction to be appended, we test the following options:
\begin{quote}
\begin{small}
{\color{black}
{\fontfamily{cmvtt}\selectfont
1: "{Generate a question about the details of an object in the image}"\\
2: \gn{"Generate a question about the details of an unusual object in the image"}\\
3: \gn{"Generate a question about the color of a small portion of the image"}\\
4: \gn{"Generate a question about a countable object quantity in the image"}\\
5: "{Generate a question about an unusual countable object quantity in the image}"\\
6: "{Generate a question about the vibe of the image}"\\
7: "{Generate a question about the artistic style of the image}"\\
8: "{Generate a question about the presence or not of an unusual object}"
}}
\end{small}
\end{quote}
\noindent\\
We utilize InternVL3-2B as an evaluator to identify challenging categories by assessing the accuracy on a subset of our LoCoMo validation split.
This leads to the identification of three particularly challenging categories: types 2, 3, and 4 (in \gn{green}). Based on these findings, we employ InternVL3-78B to generate VQA questions from these challenging categories across the entire LoCoMo dataset, maintaining our established training, validation, and test splits, while removing the few images presented in multiple splits. These augmented splits are then used consistently with the remainder of the LoCoMo data to train, select hyperparameters for, and evaluate the MemLoRA-V adapters, respectively. We report some examples in Figure \ref{fig:vqabenchmark}.

\newpage
\subsection{Qualitative Visualization of Visual Questions and Answers}
\label{sec:supp_sub_vqaimgexamples}

\newpage

\section{Technical Details}

\subsection{Implementation Details}
Our memory system implementation follows the Mem0 framework architecture \cite{chhikara2025mem0buildingproductionreadyai}, utilizing its modular design for memory extraction, updating, and retrieval operations. For semantic memory embedding, we employ the mxbai-embed-large model (335M parameters) \cite{embed2024mxbai} accessed through Ollama \cite{ollama2023}.
The vector store is implemented using FAISS \cite{douze2025faisslibrary} with Euclidean distance as the similarity metric.
All language and vision-language models are downloaded locally from Hugging Face repositories and executed with generation temperature set to $T=0.0$ to ensure deterministic outputs during both training and evaluation.

Full-pipeline evaluation on validation and test sets processes each stage separately while following the original dataset ordering. To maintain pipeline separation and prevent information leakage between stages, memory banks containing extracted knowledge are stored independently for each experimental run. All benchmark experiments are conducted on single NVIDIA A100 GPUs with 80GB memory. The sole exception is InternVL3-78B \cite{zhu2025internvl3}, which requires distributed inference across 3-4 GPUs with tensor parallelism due to its larger parameter count.

\subsection{Model Architectures and Configurations}

Throughout our experiments, we employ language models from different families based on open-source availability, ability to follow memory instructions out-of-the-box, and overall benchmark performance on public leaderboards \cite{duan24vlmbench}. As large language models to be used as teacher we make use of GPT-OSS-120B \cite{gpt120b} (HuggingFace (HF) ID \texttt{openai/gpt-oss-120b} \cite{huggingface2025}) and the instruction-tuned version of Gemma2-27B \cite{gemmateam2024gemma2improvingopen} (HF ID \texttt{google/gemma-2-27b-it} \cite{huggingface2025}) performing competitive scores on the LMArena open leaderboard \cite{chiang2024chatbot}, while being able to run on single GPUs, helping for future development and reproducibility. Specifically, being GPT-OSS-120B the best performing model with these characteristics, we utilizie this as well as \textit{judge} for LLM-as-a-judge \cite{gu2024surveyllmasajudge} evaluations. We leave out models that would require more than one GPU to run such as Qwen2.5-78B \cite{qwen2025qwen25technicalreport}, or that were failing to perform memory operations with the default setup, such as Gemma3-27B \cite{gemmateam2025gemma3technicalreport} .
As small language models to be used as student models, we follow similar criteria testing different models that would be able to perform basic memory operations by default, and that would be small enough to fit on-device, for example below 2B parameters. We utilize the instruction-tuned version of Gemma2-2B \cite{gemmateam2024gemma2improvingopen} (HF ID \texttt{google/gemma-2-2b-it} \cite{huggingface2025}) and Qwen2.5-1.5B \cite{qwen2025qwen25technicalreport} (HF ID \texttt{Qwen/Qwen2.5-1.5B-Instruct} \cite{huggingface2025}).

As VLMs to be used as teacher, given the necessity to create a reliable benchmark, we make use of the best-performing open-source model based on \textit{Open VLM Leaderboard} \cite{duan24vlmbench}, InternVL3-78B \cite{zhu2025internvl3} (HF ID \texttt{OpenGVLab/InternVL3-78B} \cite{huggingface2025}), even though it requires at least 3 A100-80GB-GPUS to run. We utilize these teacher-generated data as ground-truth targets for training the VQA expert adapter. For student models, we employ InternVL3 variants at 1B (HF ID \texttt{OpenGVLab/InternVL3-1B} \cite{huggingface2025}) and 2B (HF ID \texttt{OpenGVLab/InternVL3-2B} \cite{huggingface2025}) parameter scales, which are able to be run in both language-only and vision-language modes easily, while performing fairly on both tasks, especially InternVL3-2B.

\subsection{Expert LoRA Adapters Training}

We apply Low-Rank Adaptation (LoRA) \cite{hu2022lora} adapters to specific components of the model architecture to enable parameter-efficient finetuning while preserving the pretrained base model. Following the original LoRA implementation and variations \cite{hu2022lora, kopiczko2024vera}, in preliminary evaluations, we find that injecting LoRA adapters to query and value projection matrices of all attention layers is enough for the \textit{extraction} and \textit{update} stages, so follow this strategy. Conversely, for generation tasks, we find that applying LoRAs on all linear layers leads to better results, so we do this instead. We keep the rank and alpha of the LoRA layers consistent, set to $r=8$ and $\alpha=16$ for all experiments. All LoRA adapters are initialized with random initialization for $A$ and zero initialization for $B$, leading to zero-adapter-initialization, ensuring the adapted model begins training from the pretrained baseline performance.

As per hyperparameter search for each adapter expert, we vary learning rate and batch size in typically used working ranges \cite{hu2022lora}. While keeping fixed the dropout to 0.1, employing AdamW optimizer with $\beta_1=0.9$, $\beta_2=0.999$, and weight decay of $0.01$. The maximum number of training epochs is set to $E=50$ for all stages, but due to early stopping in most cases the actual number of epochs is significantly lower. Only the \textit{update} stage typically requires a larger number of epochs, while the \textit{generation} stage typically requires the least amount.
All training and evaluations are performed using brain floating point half-precision (BF16) \cite{kalamkar2019studybfloat16deeplearning} to reduce memory footprint and accelerate training and inference speed.

\subsection{Evaluation Configuration}

For our composite metric \textit{L}, we compute ROUGE-1 \cite{lin04rouge}, METEOR \cite{BanerjeeL05meteor}, BERTScore-F1 \cite{Zhang20bertscore} and SentenceBERT \cite{reimers2019sentencebert} using implementations in the Mem0 codebase \cite{chhikara2025mem0buildingproductionreadyai}. The LLM-as-a-judge evaluation uses GPT-OSS-120B with temperature $T=0.0$ for reproducibility over time, while as prompt we utilize the same as used by Mem0, which is reported below.
\begin{quote}
\begin{small}
{\color{black}
{\fontfamily{cmvtt}\selectfont
Your task is to label an answer to a question as ’CORRECT’ or ’WRONG’. You will be given the following data:
    (1) a question (posed by one user to another usr), 
    (2) a ’gold’ (ground truth) answer, 
    (3) a generated answer
which you will score as CORRECT/WRONG.

The point of the question is to ask about something one user should know about the other user based on their prior conversations.
The gold answer will usually be a concise and short answer that includes the referenced topic, for example:
Question: Do you remember what I got the last time I went to Hawaii?
Gold answer: A shell necklace
The generated answer might be much longer, but you should be generous with your grading - as long as it touches on the same topic as the gold answer, it should be counted as CORRECT. 

For time related questions, the gold answer will be a specific date, month, year, etc. The generated answer might be much longer or use relative time references (like "last Tuesday" or "next month"), but you should be generous with your grading - as long as it refers to the same date or time period as the gold answer, it should be counted as CORRECT. Even if the format differs (e.g., "May 7th" vs "7 May"), consider it CORRECT if it's the same date.

Now it's time for the real question:
Question: \gr{\{question\}}
Gold answer: \gr{\{gold\_answer\}}
Generated answer: \gr{\{generated\_answer\}}

First, provide a short (one sentence) explanation of your reasoning, then finish with CORRECT or WRONG. 
Do NOT include both CORRECT and WRONG in your response, or it will break the evaluation script.

Just return the label CORRECT or WRONG in a json format with the key as "label".
}}
\end{small}
\end{quote}

VQA accuracy is computed as exact string match after lowercasing and removing leading/trailing whitespace.

\newpage
\section{Additional Experiments}

\subsection{VLM performance at different scales}

To evaluate the scalability of our approach, we assess the performance of InternVL3 models across various parameter scales on the LoCoMo benchmark and our introduced VQA task. Table~\ref{tab:vlmbench2} presents results for models ranging from 1B to 78B parameters, evaluated using the composite metric (\textit{L}), LLM-as-a-judge (\textit{J}), and VQA accuracy (\textit{V}). While larger models (8B and above) demonstrate progressively stronger performance—with InternVL3-78B achieving 92.0\% VQA accuracy—these models do not fit on device, even requiring distributed inference across at least 3 GPUs for InternVL3-78B. 

Notably, our MemLoRA-V expert adapters trained on smaller base models achieve competitive results: InternVL3-2B\textit{+Exp} attains the highest \textit{L} (44.6) and \textit{J} (40.3) scores among all tested configurations up to 38B parameters, while reaching 81.3\% VQA accuracy, higher than 8B model. This demonstrates that specialized adapters can substantially enhance smaller models' memory capabilities, approaching the performance of models several times larger.

\begin{table}[H]
\centering
\caption{\textbf{Evaluation on LoCoMo benchmark and newly introduced VQA task of Mem0-V using InternVL3 model family}. Evaluation done in terms of our composite metric (\textit{L}), LLM-as-a-judge (\textit{J}), and accuracy in our VQA task (\textit{V}).}
\vspace{-0.2cm}
\label{tab:vlmbench2}
\begin{tabular}{cccc}
\toprule
\textit{VLM}& \textit{L} & \textit{J} & \textit{V} \\
\midrule
InternVL3-1B & 13.7 & 9.0 & 50.0 \\ 
InternVL3-2B & 32.2 & 27.0 & 70.8 \\
InternVL3-8B & 25.1 & 29.2 & 74.0 \\
InternVL3-38B & 34.3 & 35.6 & \underline{87.8} \\
InternVL3-78B & \underline{42.4} & \textbf{49.4} & \textbf{92.0} \\
\hline
InternVL3-1B\pr{+Exp (ours)} &29.1 & 20.2 & 69.4\\
InternVL3-2B\pr{+Exp (ours)} & \textbf{44.6} & \underline{40.3}& 81.3 \\
\bottomrule
\vspace{-0.5cm}
\end{tabular}
\end{table}

\subsection{Abilities as memory systems for LLM and VLM}

To assess the impact of visual integration on memory capabilities, we report in Table~\ref{tab:vlmbench3} the comparisons between pure language models (Qwen2.5) against vision-language models with the same language model backbone (InternVL3). For non-specialized models using Mem0, language-only models consistently outperform their multimodal counterparts on language-based LoCoMo tasks (\textit{J} and \textit{L} scores). This gap likely stems from the additional visual processing capabilities in VLMs, which may interfere with following precise memory operation instructions in text-only contexts.

However, an interesting pattern emerges when switching to MemLoRA's specialized adapters. While models with 0.5B language components maintain the previous trend favoring language-only architectures, the 1.5B models reveal a different behavior: the multimodal InternVL3-2B achieves a \textit{J} score of 40.3 compared to 36.6 for the language-only Qwen2.5-1.5B, hinting that larger VLMs may possess enhanced transfer capabilities than language-only counterparts.

\begin{table}[H]
\centering
\caption{\textbf{Evaluation on LoCoMo benchmark and newly introduced VQA task of Mem0-V using InternVL3 model family}. Evaluation done in terms of composite metric (\textit{L}), LLM-as-a-judge (\textit{J}), and accuracy in our VQA task (\textit{V}).}
\vspace{-0.2cm}
\label{tab:vlmbench3}
\begin{tabular}{cccc}
\toprule
\textit{LLM/VLM}& \textit{L} & \textit{J} & \textit{V} \\
\midrule
Qwen2.5-0.5B & 19.5 & 11.2 & - \\
InternVL3-1B & 13.7 & 9.0 & 50.0 \\
\hline
Qwen2.5-0.5B\pr{+Exp (ours)} & 28.1 & 26.6 & - \\
InternVL3-1B\pr{+Exp (ours)} &29.1 & 20.2 & 69.4\\
\hline
Qwen2.5-1.5B & 30.5 & 29.6 & - \\
InternVL3-2B & 32.2 & 27.0 & \underline{70.8} \\
\hline
Qwen2.5-1.5B\pr{+Exp (ours)} & \underline{37.3} &  \underline{36.9} & - \\
InternVL3-2B\pr{+Exp (ours)} & \textbf{44.6} & \textbf{40.3}& \textbf{81.3}\\ 
\bottomrule
\end{tabular}
\end{table}

\end{document}